\documentclass[conference]{IEEEtran}
\IEEEoverridecommandlockouts

\usepackage{cite}
\usepackage{amsmath,amssymb,amsfonts}
\usepackage{graphics}
\usepackage{graphicx}
\usepackage{textcomp}
\usepackage{xcolor}
\usepackage{subfigure} 


\usepackage[linesnumbered,ruled,vlined]{algorithm2e}
\usepackage{algpseudocode}  
\usepackage{amsmath}  

\usepackage{array}
\usepackage{multirow}
\usepackage{longtable}
\usepackage{rotating}
\usepackage{booktabs}
\usepackage{url}

\begin{document}

\title{A DeepLearning Framework for Dynamic Estimation of Origin-Destination Sequence}

\author{\IEEEauthorblockN{Zheli Xiong\IEEEauthorrefmark{2},Defu Lian\IEEEauthorrefmark{2}{*}\thanks{*Defu Lian is the corresponding author.}, Enhong Chen\IEEEauthorrefmark{2}, Gang Chen\IEEEauthorrefmark{3} and Xiaomin Cheng\IEEEauthorrefmark{2}\IEEEauthorrefmark{3}} \IEEEauthorblockA{\IEEEauthorrefmark{2}School of Data Science\\ University of Science and Technology of China, Hefei, China\\ Email: \{liandefu,cheneh,wh5606\}@ustc.edu.cn}\{zlxiong\}@mail.ustc.edu.cn\IEEEauthorblockA{\IEEEauthorrefmark{3}Yangtze River Delta Information Intelligence Innovation Research Institute, China\\ Email: cheng@ustc.win}}

\maketitle

\begin{abstract}
OD matrix estimation is a critical problem in the transportation domain. The principle method uses the trafﬁc sensor measured information such as trafﬁc counts to estimate the trafﬁc demand represented by the OD matrix. The problem is divided into two categories: static OD matrix estimation and dynamic OD matrices sequence(OD sequence for short) estimation. The above two face the underdetermination problem caused by abundant estimated parameters and insufﬁcient constraint information. In addition, OD sequence estimation also faces the lag challenge: due to different traffic conditions such as congestion, identical vehicle will appear on different road sections during the same observation period, resulting in identical OD demands correspond to different trips. To this end, this paper proposes an integrated method, which uses deep learning methods to infer the structure of OD sequence and uses structural constraints to guide traditional numerical optimization. Our experiments show that the neural network(NN) can effectively infer the structure of the OD sequence and provide practical constraints for numerical optimization to obtain better results. Moreover, the experiments show that provided structural information contains not only constraints on the spatial structure of OD matrices but also provides constraints on the temporal structure of OD sequence, which solve the effect of the lagging problem well.
\end{abstract}


\section{Introduction}
With the development of big traffic data, a large amount of traffic data has been widely used in traffic applications such as route planning, flow prediction and traffic light control. Traffic demand describes the trips between the divided areas, commonly referred to as the OD (Origin-Destination) matrix. The OD matrix has significant value for various traffic tasks such as trafﬁc ﬂow prediction\cite{li2017diffusion}, trajectory prediction\cite{altche2017lstm} and location recommendation\cite{lian2020geography}. On the one hand, the change in traffic demand between regions will affect the flow of the road sections, leading to a change in the optimal vehicle path. On the other hand, OD estimation will lead to more effective traffic regulation since the already known OD demants can provide rationality for management strategies. However, traffic demand is the data that sensors cannot directly observe, and it must be obtained through other traffic data, such as the traffic flow, to estimate the matrix.

OD matrix estimation is mainly divided into two categories: static OD matrix estimation\cite{behara2020novel} and dynamic OD sequence estimation\cite{cascetta1993dynamic}. Static OD estimation uses integrated traffic counts to estimate the total traffic demand in the period. In comparison, dynamic OD sequence estimation uses time-varying traffic counts in a sequence of intervals to estimate the corresponding OD matrix sequence.
The bi-level framework is commonly used for OD estimation\cite{bert2009dynamic}, which is divided into upper and lower levels. The upper level adjusts the OD matrices by minimizing the numerical gap between real and estimated trafﬁc counts by solving a least squares problem. The optimizing method is mainly gradient-based\cite{spiess1990gradient} and optimizes the OD matrices by the steepest descent method. The lower level assigns traffic demand to road sections by analysis\cite{maher2001bi} or simulation\cite{behara2020novel} method. The upper and lower layers updates iteratively and converge to an optimum point.

Due to the limited number of road sections and abundant parameters to be estimated in the OD matrices, the optimization problem is heavily underdetermined\cite{robillard1975estimating}. Some researchers have alleviated this problem by adding other observable information, such as travel speed\cite{jaume2015integrated}, cellular probe\cite{calabrese2011estimating}, and bluetooth probe\cite{behara2020novel}to the estimator. In addition to the above challenge, dynamic OD sequence estimation faces another lag challenge: due to different traffic conditions such as congestion, identical vehicle will appear on different road sections during the same observation period, resulting in identical OD demands correspond to different trips\cite{tavana2001internally}, so current OD matrix will refer to different time-varying traffic counts according to different traffic conditions. Furthermore, traffic conditions are caused by OD marices before and after the current OD matrix, which causes a temporal relationship between these OD matrices. To this end, some studies propose fixed maximum lags\cite{cheng2022real}, which assume that vehicles can complete their trip within a fixed maximum interval amount to eliminate the influence of observation intervals amount. And others like \cite{tavana2001internally} propose using the temporal overlap in consecutive estimation intervals to alleviate the influence between OD matrices, as shown in Fig. 1.

To address these challenges above, we use deep learning to fit the mapping relationship between the time-varying traffic counts and the structure of the OD sequence to learn the impact of lagged traffic counts on the OD structure and the relationship between OD sequences.

Consider that numerical space makes the learning space too large for deep learning. However, distribution can constrain the learning space's size and reflect the structural information. Furthermore, simply using NN to infer numerical matrices will lose important information about assignments presented in the bi-level framework. Therefore, we propose a method that integrates deep learning and numerical optimization, which merges the inferred structural information into the upper-level estimator of the bi-level framework. Providing the spatial and temporal constraints for optimization can effectively avoid to fall into local optimum in advance and help to optimize to a better result.

In this paper, we further deliver the following contributions.
\begin{itemize}

\item We proposed a deep learning model to infer OD sequence structure which extracts the spatio-temporal constraints from time-varying traffic counts effectively.
\item We novelly integrates deep learning method and  bi-level framework to solve OD sequences estimation.
\item Through our experiments, it is verified that the structural knowledge inferred by deep learning can provide great help for numerical optimization.
\end{itemize}

\begin{figure}[htbp]
\centerline{\includegraphics[width=8cm,height=4cm]{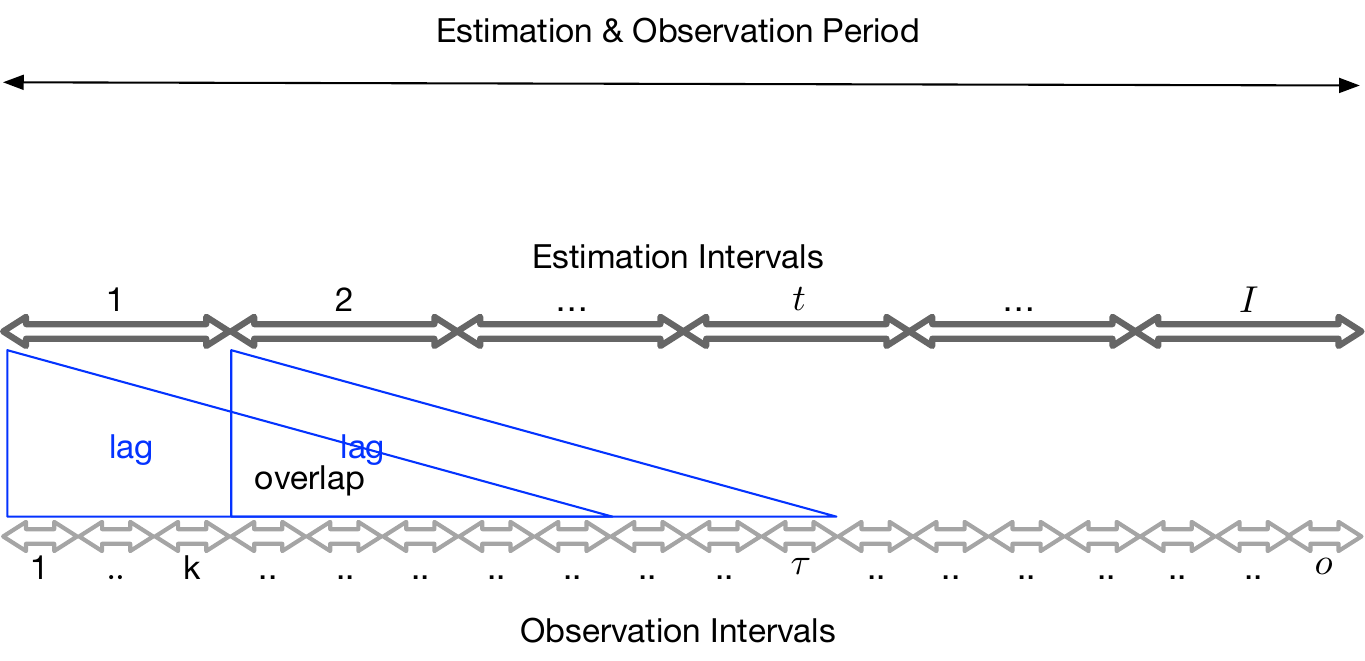}}
\caption{Illustration of intervals}
\label{fig}
\end{figure}

\noindent\textbf{Reproducibility.} Code and data are available at: \url{https://github.com/shaun19920309/A-DeepLearning-Framework-for-\\Dynamic-Estimation-of-Origin-Destination-Sequence}.

\section{Related Works}
OD matrix estimation can be mainly divided into static OD matrix estimation and dynamic OD matrices estimation. 
\subsection{static OD matrix estimation}\label{AA}
For solving the static OD matrix estimation, gravity mode adopts a "gravitational behavior" for trip demand and builds a linear or nonlinear regression model\cite{low1972new}. The maximum likelihood technique\cite{spiess1987maximum} estimates the OD matrix over the sampled matrix on the presumption that OD pairs follow independent Poisson distributions. The entropy maximizing/information minimization method\cite{van1980most} seeks to select an estimated OD matrix that adds as little information as possible from traffic counts in order to match the underdetermined problem. The Bayesian method\cite{maher1983inferences} additionally resolves OD matrices by maximizing the posterior probability, which utilizes a mix of the prior OD matrix and the observations. The maximizing/information minimization approach is a particular instance of Bayesian method when the prior is given only a minimal amount of confidence. A method that explicitly considers both observed flow biases and the target OD matrix is built using generalized least squares (GLS)\cite{cascetta1984estimation}. All of the aforementioned techniques need to use the prior OD matrix, which may be out of date and lead to estimation bias. Additionally, travel time\cite{barcelo2010travel}, travel speed\cite{jaume2015integrated}, and turning proportions\cite{alibabai2008dynamic} are also employed directly in OD estimates due to the availability of massive traffic data and traffic simulation.

\subsection{dynamic OD matrices estimation}\label{AA}
Static OD matrix estimation can only estimate OD for a specific period. While a period is divided into multiple intervals, researchers propose dynamic OD matrix estimation, which aims to estimate the OD matrices in the corresponding time intervals and estimate the entire OD sequence further by using road traffic and other information. However, as mentioned above, the lagged flow may vary in successive intervals according to different traffic conditions, so the estimation by simply applying the static OD matrix is no longer applicable. Dynamic OD matrix estimation is divided into two types: off-line case and on-line case\cite{bert2009dynamic}.

\subsubsection{Off-line case}
The off-line method mainly discusses the direct estimation of the OD sequence when only given the corresponding observation sequence. \cite{cascetta1993dynamic} discusses off-line estimation methods, such as the simultaneous and sequential method, and further\cite{spiess1990modelling} proposes a correction method based on the average OD matrix. The simultaneous method focuses on establishing an estimator to optimize all OD matrix slices simultaneously, its idea is similar to the static OD matrix estimation. Compared to the simultaneous method, the sequential method only estimates one matrix slice at a time and infers the next OD matrix slice based on the historical OD matrices that have been estimated. The average-based correction method estimates an average OD over the observation period, then estimates coefficients, which are multiplied by the average OD to obtain the final OD sequence. Since each interval corresponds to an OD matrix slice, there are large amounts of parameters with non-negative constraints to be estimated. The main optimization methods proposed by researchers are the gradient projection method\cite{lundgren2008heuristic} and the Simultaneous Perturbation Stochastic Approximation(SPSA)\cite{antoniou2014framework}. Similar to static OD estimation, some researchers have integrated various traffic observations, such as vehicle transit time, travel speed, etc., into the estimators\cite{bullejos2014due} to improve the accuracy.

\subsubsection{On-line case}

The on-line method has been widely studied. Unlike the off-line method, it requires the historical periodic OD sequence and the current observation. Traditional modeling methods include Kalman filter CF\cite{bishop2001introduction}\cite{van1994improved}, LSQR algorithm\cite{bierlaire2004efficient}, Canonical Polyadic (CP) decomposition\cite{ren2017efficient}. \cite{liu2020dynamic} used principal component analysis (PCA) combine several machine learning models, and \cite{schmid2010dynamic}\cite{cheng2022real} utilize Dynamic Mode Decomposition (DMD) based method. At the same time, given the excellent performance of deep learning in prediction, some researchers use deep learning methods for the dynamic prediction of OD sequences. For example, \cite{xiong2020dynamic} used a Graph Neural Network(GNN) to capture the spatial topology information of the graph structure and combined it with the traditional CF algorithm to improve the accuracy of the OD matrix prediction. \cite{chu2019deep}\cite{pan2019urban} used a Recurrent Neural Network(RNN) to capture the temporal features of prior OD sequence evolution to predict the current OD matrix. The primary relationship is that the off-line methods can be used to provide a better initialization for on-line methods\cite{barcelo2014practical}.

\begin{table}
\centering
	\caption{Illustration of notations}
	\label{table}
	\setlength{\tabcolsep}{3pt}
	\renewcommand\arraystretch{1}
	\begin{tabular}{m{1cm}<{} m{7cm}<{\centering}}
	\toprule
	
		$n_{od}$ & the number of OD nodes \\
		$n_{sec}$ & the number of road sections\\
	       $I$  & the number of estimation intervals \\
	       $o$ & the number of observation intervals \\
	       $\pmb{\epsilon}_{\tau}$ & a vector of traffic counts during observation interval $\tau$, $\pmb{\epsilon}_{\tau} \in \mathbb{R}^{n_{sec}}$ \\
	       $\pmb{E}$ &   a tensor composed of $\pmb{\epsilon}_{1}$ to $\pmb{\epsilon}_{o}$, $\pmb{E} \in \mathbb{R}^{o \times n_{sec}}$\\
	       	$\pmb{n}_{i}$  &  a vector represents the OD node $i$, $\pmb{n}_{i}\in \{-1,0,1\}^{n_{sec}}$ \\
		$\pmb{N}$ &   a tensor composed of $\pmb{n}_{1}$ to $\pmb{n}_{n_{od}}$, $\pmb{N} \in \{-1,0,1\}^{n_{od} \times n_{sec}}$\\
		$M_{ijt}$  & the number of traffic trips from $\pmb{n}_{i}$ to $\pmb{n}_{j}$ during estimation interval $t$ \\
		$\pmb{T}$   &  a tensor transformed from OD sequence, $\pmb{T} \in \mathbb{R}^{I \times  n_{od}^{2}}$\\
		$\tilde{\pmb{T}}$   &  a initial OD sequence is as a starting point  of optimization.\\
		$\hat{\pmb{T}}$   &  an optimized OD sequence during optimization phase.\\
		
		$\tilde{\pmb{p}}_{t}$   & production flow, a vector of trips leaving from each node during estimation interval $t$, $\tilde{\pmb{p}} \in \mathbb{R}^{n_{od}}$  \\
		$\tilde{\pmb{a}}_{t}$   &  attraction flow, a vector of trips arriving at each node during estimation interval $t$, $\tilde{\pmb{a}} \in \mathbb{R}^{n_{od}}$  \\
		$\pmb{p}$  & global production flow, a vector concatenated from $\tilde{\pmb{p}}_{1}$ to $\tilde{\pmb{p}}_{I}$, $\pmb{p} \in \mathbb{R}^{I \cdot n_{od}}$ \\
		$\pmb{a}$  & global attraction flow, a vector concatenated from $\tilde{\pmb{a}}_{1}$ to $\tilde{\pmb{a}}_{I}$, $\pmb{a} \in \mathbb{R}^{I \cdot n_{od}}$ \\
		$\pmb{D}_{E}$   &  a tensor of the distribution of traffic counts by normalizing $\pmb{E}$\\
		$\pmb{d}_{p\ or \ a}$   &  a vector of the distribution by normalizing $\pmb{p}$ or $\pmb{a}$ \\
		$\bar{\pmb{d}}_{p\ or \ a}$   &  the inferred distribution of production flow or attraction flow from deep learning model\\
		$\pmb{d}_{p\ or \ a}^{*}$   &  the inferred distribution of production flow or attraction flow from the best trained deep learning model during inference phase \\
		$\hat{\pmb{d}}_{p\ or \ a}$   &  the optimized distribution of production flow or attraction flow from bi-level framewrok during optimization phase \\
		\bottomrule
	\end{tabular}
	\label{tab1}
\end{table}

\section{Preliminary}
\subsection{Definitions}\label{AA}

As shown in Table 1, an OD node is a cluster created by grouping the intersections of road sections in the city network, and the roads connect the same OD pair are aggregated to one road section (see Fig. 3). Considering a city network consists of $n_{od}$ OD nodes and $n_{sec}$ road sections. We use a vector of length $n_{sec}$ to represent OD node $\pmb{n}_{i}$, with $n_{ij}=1$ if road section $j$ enters node $\pmb{n}_{i}$, -1 if it exits from $\pmb{n}_{i}$, and 0 if it is not connected to $\pmb{n}_{i}$.

During an estimation period divided into $I$ equal intervals $t = 1, 2, 3, . . ., I$, and an obsevation period divided into $o$ equal intervals $\tau = 1,2,3,...,o$.  $\pmb{\epsilon}_{\tau}$ denotes the traffic counts of all road sections during observation interval $\tau$. The traffic trips from $\pmb{n}_{i}$ to $\pmb{n}_{j}$ during estimation interval $t$ is denoted by $M_{ijt}$. In addition, we transform the OD sequence into a tensor denoted by  $\pmb{T} \in \mathbb{R}^{1\times I \times  n_{od}^{2} \times 1}$, and $\pmb{\varepsilon}$ denotes a vector concatenated from $\pmb{\epsilon}_{1}$ to $\pmb{\epsilon}_{o}$. For $\varepsilon_i \in \pmb{\varepsilon}$, we normalize its traffic count on each road section with respect to all observation intervals as $D_{E_{ij}} = \frac{E_{ij}}{\sum\limits_{i}^{o} \sum\limits_{j}^{n_{sec}} E_{ij}}$.

Production flows $\tilde{\pmb{p}}_{t}$ is used to represent a vector records the number of trips leaving each node during estimation interval $t$ as $\tilde{\pmb{p}}_{t}=\sum\limits_{j}^{n_{od}}M_{ijt}$. And attraction flows $\tilde{\pmb{a}}_{t}$ is to used to represent a vector records the number of trips that arrive at each node during estimation interval $t$ as $\tilde{\pmb{a}}_{t}=\sum\limits_{i}^{n_{od}}M_{ijt}$.

Additionally, the global production flows $\pmb{p}$ is a vector concatenated from $\tilde{\pmb{p}}_{1}$ to $\tilde{\pmb{p}}_{I}$, and the global attraction flows $\pmb{a}$ is a vector concatenated from $\tilde{\pmb{a}}_{1}$ to $\tilde{\pmb{a}}_{I}$, which are used to denotes the production flows and attraction flows of OD sequence, respectively. Moreover, we normalize their flow on each OD node with respect to all OD nodes and  all estimation intervals as $d_{pi}=\frac{p_{i}}{\sum\limits_{k}^{I \times n_{od}} {p}_{k}}$ and $d_{ai}=\frac{a_{i}}{\sum\limits_{k}^{I \times n_{od}} {a}_{k}}$, respectively.

In the inference phase, $\bar{\pmb{d}}_{p\ or \ a}$ denotes the inferred distribution of production flow or attraction flow from deep learning model when being trained. And the best inferred distribution is denoted by $\pmb{d}_{p\ or \ a}^{*}$.

In the optimization phase, $\hat{\pmb{T}}$ denotes the tensor of optimized OD sequence from bi-level framewrok at each iteration, and correspondingly, $\hat{\pmb{d}}_{p\ or \ a}$ denotes the optimized distribution of production flows or attraction flows.

\subsection{Bi-level framework}\label{AA}
In the bi-level framework, the estimation will start from an initialized OD matrix $\tilde{\pmb{T}}$. In the lower level, for each observation interval, trips in every OD pair are allocated to the road sections in an analytical or simulative way, and then an allocation tensor $\pmb{P}$ is obtained. $\pmb{P}_{\tau t}$ represents a matrix of proportion that OD matrix $\pmb{T}_{t}$ allocated to road sections during the observation interval $\tau$, and $\pmb{P}_{\tau t} \pmb{T}_{t}$ indicates the corresponding traffic counts. In the upper layer, the traffic counts assigned by $\pmb{T}_{1},\pmb{T}_{1},...,\pmb{T}_{t}$ are summed to get the traffic counts $\pmb{\epsilon}_{\tau}$ as shown in Fig 1. Optimizing the least squares estimator reduces the gap between optimized traffic counts $\hat{\pmb{\epsilon}}_{\tau}$ and real traffic counts $\pmb{\epsilon}_{\tau}$ throughout the whole observation period $\tau=1,2,...,o$ to find a better estimated OD sequence. By iteration repeats the alternation of upper and lower levels, the final estimated OD matrix sequence is obtained when converges. The least squares estimator is formulated as follows:
\begin{IEEEeqnarray}{c} 
Z(\hat{\pmb{T}})= \min_{\hat{\pmb{T}}_{1} \ge 0,...,\hat{\pmb{T}}_{I} \ge{0}}  \sum_{\tau=1}^{o}   \frac{1}{2}(\pmb{\epsilon}_{\tau}-\hat{\pmb{\epsilon}}_{\tau})^{\mathrm{T}}(\pmb{\epsilon}_{\tau}-\hat{\pmb{\epsilon}}_{\tau}) \IEEEnonumber\\  
where\  \hat{\pmb{\epsilon}}_{\tau}=   \sum_{k=1}^{t}  \pmb{P}_{\tau k} \hat{\pmb{T}}_{k}\IEEEnonumber\\ 
\end{IEEEeqnarray}

The assignment tensor $\pmb{P} \in  \mathbb{R}^{  o   \times I   \times n_{sec} \times n_{od}^{2}}$ , can be derived by analysis, for example by taking into account stochastic user equilibrium on traffic counts or by simulation using a simulator like SUMO\cite{lopez2018microscopic}. It displays the ratio of each OD trip to each road section during estimation intervals. Similar to\cite{behara2020novel}, our assignment tensor P is computed using a back-calculation technique based on traffic counts generated by the simulator during each iteration:

\begin{displaymath}
\pmb{P}=
	\left( \begin{array}{ccc}
	\pmb{P}_{11} & \ldots & \pmb{P}_{1I}\\
	\pmb{P}_{21} & \ldots & \pmb{P}_{2I}\\
	\vdots &  \ddots &\\
	\pmb{P}_{o1} & \dots &
	\left( \begin{array}{ccc}
	p_{11}  & \ldots p_{1\ n_{od}^{2}} \\
	p_{21}  & \ldots p_{2\ n_{od}^{2}}\\
	\vdots &  \vdots \\
	p_{n_{sec}\ 1} & \ldots p_{n_{sec} \ n_{od}^{2}}\\
	\end{array} \right) _{oI}
	\end{array} \right)
\end{displaymath}

\subsection{Optimization}\label{AA}
Considering Eq(1) is a problem with abundant non-negativity constraints on $\pmb{T}$, the gradient projection method is commonly used\cite{lundgren2008heuristic}. The idea is to evaluate the directional derivatives of the objective function at the current point, and to obtain a descent direction by making a projection on the non-negativity constraints.

It is worth noting that we adjust the update step $\pmb{T}^{k+1}:=\pmb{T}^{k} \oplus (\lambda^{k} \odot \pmb{d}^{k})$ to $\pmb{T}^{k+1}:=\pmb{T}^{k} \odot (\pmb{e}+\lambda^{k} \odot \pmb{d}^{k})$. It has been proved that compared with the ordinary update step, this adjustment significantly improves the optimization speed of OD matrices\cite{spiess1990gradient}, since it proposed that update steps for larger variables should be greater. For the upper bound of step size $\lambda^{k}_{max}$ at itertion $k$, we give the corresponding adjustment  $\lambda^{k}_{max}=\mathop{\min}\{\frac{-1}{\pmb{d}^k_i}|\forall i:\pmb{d}^k_i<0\}$. Since at the $(k+1)^{th}$ iteration,  $\mathbf{A}_{2}\pmb{T}^{k} > 0$ and ensure $\mathbf{A}_{2}\pmb{T}^{k+1} > 0$, let $\mathbf{A}_{2}\pmb{T}^{k} \odot (\pmb{e}+\lambda^{k} \odot \pmb{d}^{k}) > 0$, which implies $\pmb{e}+ \lambda^{k} \odot \pmb{d}^{k}>0$. Therefore, $\lambda^{k}<\frac{-1}{\pmb{d}^{k}_{i}},\forall i:\pmb{d}^k_i<0$, and then we have $\lambda^{k}_{max}=\min\{\frac{-1}{\pmb{d}^{k}_{i}}\},\forall i:\pmb{d}^k_i<0$.

Finally, we search for the optimal step size $\lambda^{*k}$ at iteration $k$ based on Eq(2) and then determine the executable step size $\lambda^{k}$ according to $\lambda^{*k}$ and $\lambda^{k}_{max}$, that is, if $\lambda^{*k}<\lambda^{k}_{max}$, set $\lambda^{k}=\lambda^{*k}$; otherwies, set $\lambda^{k}=\lambda^{k}_{max}$.

\begin{equation}
\min_{\lambda^{k}}Z(\hat{\pmb{T}}^{k} \odot (\pmb{e}+\lambda^{k} \odot \pmb{d}^k)) 
\end{equation}

Where $\odot$ denotes the element-wise product and $\pmb{e}$  is a tensor of 1s with the same dimension as $\hat{\pmb{T}}^{k}$.

\begin{algorithm}
\caption{Imporved Gradient Projection}  
step 0: Giving an initial point that satisfies the constraints $\pmb{T}^0$, expand $\pmb{T}^{0}$ into a vector as $F(\pmb{T}^{0}) \in \mathbb{R}^{I \cdot n_{od}^{2}}$, let $k=0$, threshold $\epsilon>0$;

step 1: Construct search direction at $\pmb{T}^{k}$. Let $\mathbf{A}=\left[\begin{matrix} \mathbf{A}_{1}\\  \mathbf{A}_{2} \end{matrix}\right]$, $b=\left[\begin{matrix} 0\\ 0\end{matrix}  \right]$, $\mathbf{A}_{1}F(\pmb{T}^{k})=0, \mathbf{A}_{2}F(\pmb{T}^{k})>0$. 

step 2: Let $\mathbf{M}=\left[\begin{matrix} \mathbf{A}_{1}\\  \mathbf{E} \end{matrix} \right]$,  Let $\mathbf{P_{M}}=\mathbf{I}$ if $\mathbf{M}$ is empty, and $\mathbf{P_{M}}=\mathbf{I}-\mathbf{M}^{\mathrm{T}}(\mathbf{M}\mathbf{M}^{\mathrm{T}})^{-1}\mathbf{M}$ otherwise.

step 3: Calculate  $\pmb{d}^k=-\mathbf{P_{M}}\nabla Z(\pmb{T}^k)$. If $\|\pmb{d}^k\| \neq 0$, to $step 5$; otherwise goto $step 4$.

step 4: Calculate $\left[\begin{matrix} \lambda\\  \mu \end{matrix}\right]=(\mathbf{MM^{\mathrm{T}}})^{-1}\mathbf{M}\nabla Z(\pmb{T}^k)$. If $u\geq 0$ stop and $\pmb{T}^k$ is KKT point. Otherwise let $u_{i0}=\mathop{\min}\{u_i\}$, and remove the row corresponding to $u_{i0}$ from $\mathbf{M}$ and goto $step 2$.

step 5: Determine the step size. Let $\lambda^{k}_{max}=\mathop{\min}\{\frac{-1}{\pmb{d}^k_i}|\forall i:\pmb{d}^k_i<0\}$, and determine $\lambda^{*k}$ based on Eq(2). If $\lambda^{*k}<\lambda^{k}_{max}$, let $\lambda^{k}=\lambda^{*k}$; otherwies, let $\lambda^{k}=\lambda^{k}_{max}$.

step 6: Let $\pmb{T}^{k+1}:=\pmb{T}^{k} \odot (\pmb{e}+\lambda^{k} \odot \pmb{d}^{k}), k:=k+1$, goto $step 2$.
\end{algorithm}

\section{method}
The pipline of our proposed method will be described in detail in this section, including sampling the probe flow to compose datasets, training and inference of NN models, and combining inferred spatial-temporal structural distributions into numerical optimization.

\subsection{Probe Trafﬁc Sampling}\label{AA}
Firstly, as presented in our previous work on static OD estimation, most of the important trips are in a small part of OD pairs, leat to the values of other OD pairs are relatively small. So the production and attraction ﬂows of an OD matrix will be uneven in the reality, and this property implies the structural information of the OD matrix. Moreover, as we mentioned in Part 1, we infer distributions rather than real numbers to reﬂect the structure, so the exact value is not a concern. Therefore, it is feasible to set a sparse matrix with limited values of non-zero elements to reconstruct the specific structure information of an real OD matrix. To this end, we set $m$ to represent the maximum value of OD pairs and make it relatively small to speed up the sampling process.

Secondly, since the traffic congestion will cause lag problem, we need some probe vehicles to explore the traffic congestion. We form a original matrices sequence(OMS) by collecting $m$ vehicles from these important OD pairs (with the number of trips $>m$) of the real OD sequence as shown in the left part of Fig. 2, these trips can be a combination of various data, such as car-hailing service data and GPS data since these vehicles can all be seen as probe vehicles. Then, we resample each OD paris on a scale of 0.0-1.0 from OMS to obtain a dataset composed of generated OD sequences, and calculate the corresponding global distributions $\pmb{d}_{p}$, $\pmb{d}_{a}$ and $\pmb{D}_{E}$. The advantage of doing so is that, although we sample from a small number $m$ of vehicles, it also can reconstruct the relationship between various traffic counts distribution and its corresponding golobal distributions under the real traffic conditions. 

Finally, $\pmb{D}_{E}$ are used as inputs of NN, $\pmb{d}_{p}$ or $\pmb{d}_{a}$ are used as labels, we form the dataset $(\pmb{D}_{E},\pmb{d}_{p})$ and $(\pmb{D}_{E},\pmb{d}_{a})$ and train the two models separately.

\subsection{Dynamic Distribution Inference}\label{AA}

The spatio-temporal evolution of traffic counts can effectively characterize the OD sequence. We utilize traffic counts of observation interval $t$ to $t \times k+ \delta$ (for $t \in [0,I-1]$, $k=\frac{o}{I}$, $\delta>0$)  as input to characterize the OD matrix of estimation interval $t$. If $\delta>k$, the observations overlap between two estimation intervals as shown in Fig 1, it indicates that the OD matrix of the current estimation interval $t$ will affect the traffic counts of the following $t \times k+\delta$ observation intervals due to the lag problem.

In order to obtain the global distributions, we need to consider the mutual influence relationship of each OD node in spatial and temporal. For example, if the trips of node $n_{i}$ at observation interval $t \times k+\delta_{1}$ and node $n_{j}$ at $t \times k+\delta_{2}$ both need to pass through road section $e$ at $t \times k+\delta$ when the traffic count has been given($\delta >\delta{1}$; $\delta>\delta_{2}$), so there will be pairwise spatial-temporal dependencies between OD nodes.

\begin{figure*}[htbp]
\centerline{\includegraphics[width=18cm,height=6cm]{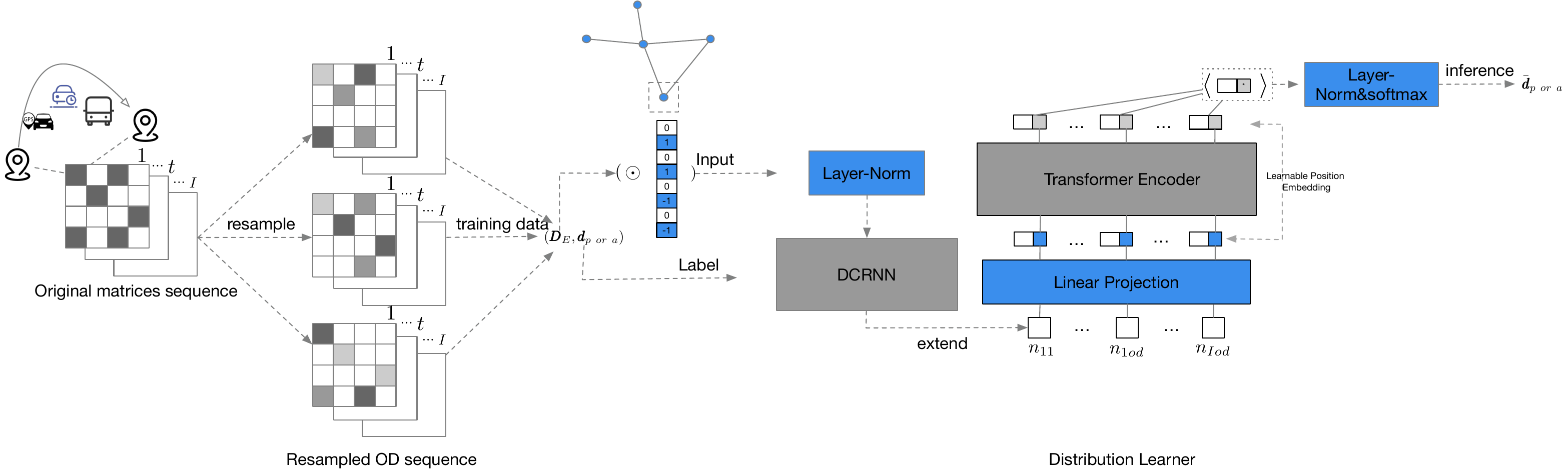}}
\caption{Framework: Data Sampling \& Distribution Learner}
\label{fig}
\end{figure*}

\subsubsection{DCGRU} 
In deep learning field, many studies have shown that the GNN+RNN based method can extract the spatio-temporal features well\cite{pareja2020evolvegcn}. Therefore, we choose the DCGRU model as the feature extractor of each OD matrix. It combines Diffusion Convolutional Network(DCN, a spatial-based GNN model) and Gated Recurrent Unit(GRU,an improved RNN model) and has been studied to have an outstanding performance in capturing the long term spatio-temporal evolution characteristics of traffic flow\cite{li2017diffusion} .

DCN can be adapted to deal with dependency between objects in non-Euclidean spaces according to node features $\pmb{\chi}$ and the adjacency matrix $\pmb{W}$. The $K$-step graph diffusion convolution is calculated to extract the upstream and downstream spatial dependencies between their surrounding $K$-order neighbor nodes and form an integrated graph signal, which is formulated as below:

\begin{IEEEeqnarray}{c}
\pmb{\chi}_{:,e\star g}=\sum_{k=0}^{K-1}(\theta_{k,1}(\pmb{D}_{o}^{-1}\pmb{W})^{k}+\theta_{k,2}(\pmb{D}_{I}^{-1}\pmb{W}^{T})^{k})\pmb{\chi}_{:,e}\IEEEnonumber\\
for \ e\  \in \{1,...,n_{sec} \}
\end{IEEEeqnarray}

$\pmb{\chi}_{\star g}$ is the graph signal obtained after each OD node fuses the $K$ order neighbors in every dimension $e$. $\pmb{\theta} \in \mathbb{R}^{K \times 2}$ are learnable parameters for the filter. $\pmb{D}_{o}$ represents the diagonal matrix of the out-degree matrix of graph $g$, and $\pmb{D}_{I}$ represents the diagonal matrix of the in-degree matrix, $\pmb{D}_{o}^{-1}\pmb{W}$, $\pmb{D}_{I}^{-1}\pmb{W}^{T}$ represent the transition matrices of the diffusion process and the reverse one respectively.

GRU sends the integrated graph signal into the cell orderly to capture the temporal dependencies. The update process of feature in the GRU cell is as follows:

\begin{IEEEeqnarray}{c} 
\pmb{r}^{(\tau)}=\sigma(\pmb{\Theta}_{r \star  g}[\pmb{\chi}^{(\tau)},\pmb{H}^{(\tau-1)}]+\pmb{b}_{r}) \\
\pmb{u}^{(\tau)}=\sigma(\pmb{\Theta}_{u \star  g}[\pmb{\chi}^{(\tau)},\pmb{H}^{(\tau-1)}]+\pmb{b}_{u}) \\
\pmb{C}^{(\tau)}=tanh(\pmb{\Theta}_{C \star  g}[\pmb{\chi}^{(\tau)},(\pmb{r}^{(\tau)} \odot \pmb{H}^{(\tau-1)}]+\pmb{b}_{c}) \\
\pmb{H}^{(\tau)}=\pmb{u}^{(\tau)} \odot \pmb{H}^{(\tau-1)}+(1-\pmb{u}^{(\tau)}) \odot \pmb{C}^{(\tau)} 
\end{IEEEeqnarray}

where $\pmb{\chi}^{(\tau)}$, $\pmb{H}^{(\tau)}$ denote the input and output at observation interval $\tau$, $\pmb{r}^{(\tau)}$, $\pmb{u}^{(\tau)}$ are reset gate and update gate at $\tau$ respectively. $\star g$ denotes the diffusion convolution defined in Eq(3) and $\pmb{\Theta}_{r}$, $\pmb{\Theta}_{u}$, $\pmb{\Theta}_{C}$ are learnable parameters for the corresponding ﬁlters.

\subsubsection{Multihead Self-Attention(MSA)}

Standard $\pmb{qkv}$ self-attention(SA) compute the attention weight $A$ over all value of elements $\pmb{v}$. $A$ is based on the of query $\pmb{q}$ and key $\pmb{k}$ of elements , and calculate the pairwise dependency between two elements of input sequence $\pmb{\zeta} \in \mathbb{R}^{(n_{od}^{2}+1) \times d}$.

Self-attention (SA)\cite{vaswani2017attention} computes the attention weight A overall value of elements $\pmb{v}$ to calculate the pairwise dependency between two elements of input sequence  $\pmb{\zeta} \in \mathbb{R}^{(I \cdot n_{od}^{2}) \times d}$ where $A$ is based on the query $\pmb{q}$ and key $\pmb{k}$ of elements.

\begin{equation}
[\pmb{q},\pmb{k},\pmb{v}]=\pmb{\zeta}\pmb{U}_{qkv},\  \pmb{U}_{qkv} \in \mathbb{R}^{d \times 3d_{h}}
\end{equation}
\begin{equation}
A=softmax(\pmb{qk}^\mathrm{T} / \sqrt{d_{h}})
\end{equation}
\begin{equation}
SA(\pmb{\zeta})=A\pmb{v}
\end{equation}

We projected concatenated outputs from MSA, which runs $h$ SA procedures concurrently. The dimensions are kept constant by setting $d_{h}$ to $d/h$, where $h$ is the number of heads.

\begin{IEEEeqnarray}{c} 
MSA(\pmb{\zeta})=[SA_{1}(\pmb{\zeta});SA_{2}(\pmb{\zeta});...;SA_{h}(\pmb{\zeta})]\pmb{U}_{msa}, 
\IEEEnonumber\\
\pmb{U}_{msa} \in \mathbb{R}^{h \cdot d_{h} \times d} 
\end{IEEEeqnarray}
 $\pmb{U}_{qkv}$  and $\pmb{U}_{msa}$ above are learnable parameters

\subsubsection{Distribution Learner}
We element-wise multiply the node vector $\pmb{n}_{i}$ with the distribution of traffic counts $\pmb{\epsilon}_{\tau}$ to obtain the feature of OD node $i$ at observation $\tau$. In order to get all OD nodes fetures $\pmb{\chi}$ during all the $o$ observation intervals. We expand dimention of $\pmb{N}$ to $n_{od} \times 1 \times n_{sec}$ and $\pmb{D_{E}}$ to $1 \times o \times n_{sec}$, respectively, then do the broadcast operation $\otimes$ on these two tensor as shown in Eq(12) to get the shape of $\pmb{\chi}$ as $(o, n_{od}, n_{sec})$. Then, divide by the dimension $o$ of the tensor, and take the $t \times k$ to $t \times k+\delta$ (for $t \in [0,I-1]$, $k=\frac{o}{I}$, $\delta>0$) each time to obtain the input tensor $(\delta, n_{od}, n_{sec})$. In our case, we estimate an OD sequence of 12 hours, with an estimation interval every hour and an observation interval every 10 minutes. So we have $I=12, o=72, k=6$.

\begin{equation}
\pmb{\chi}=\pmb{N} \otimes \pmb{D}_{E}
\end{equation}

Subsequently, taking $\{\pmb{\chi}^{(t \times k)},...,\pmb{\chi}^{(t \times k+\delta)} \ |\  t \in [0,I-1]\}$ as the input of the DCGRU module orderly, a hidden tensor $\pmb{H}=\{\pmb{H}^{(t \times k+\delta)} \ |\ t \in [0,I-1]\}$ is as the output with its shape is $(I, n_{od}, n_{sec})$. Then expanded $\pmb{H}$ by the dimension $I$ to obtain $(I \times n_{od}, n_{sec})$ as the input of the Transformer encoder. With a shared position embedding parameters added before and after Transformer encoder, we refer to the output as mutual vectors, there are $I \times n_{od}$ mutal vectors for each represents the spatio-temporal mutual information of corresponding OD node. Lastly, we operate element-wise addtion $\oplus$ on all these mutual vectors to one vector * containing the global infromation, and do inner production between each mutual vector and the global information vector * to give a scalar for each node, then perform the softmax operations to obtain the inferred global distribution $\bar{\pmb{d}}_{p \ or \ a}$.

For model training, we choose Jensen-Shannon Divergence(JSD) as the loss function as Eq(13), which measures the distance between two distributions symmetrically.
\begin{IEEEeqnarray}{c} 
Loss_{p\ or\ a}=JS(\bar{\pmb{d}}_{p\ or\ a}||\pmb{d}_{p\ or\ a})
\end{IEEEeqnarray}

\subsection{Estimator}\label{AA}
Like other studies in static OD estimation and the off-line case of dynamic OD sequence estimation, we adopt the bi-level framework. The difference is the least squares approach at the upper level merely seeks to reduce the gap between observed and simulated traffic counts, which only facilitates numerical similarity between estimated and real OD sequence. Therefore, we incorporate the optimization with the best inference global distributions $\pmb{d}_{p}^{*}$ and $\pmb{d}_{a}^{*}$ and choose KLD\cite{bishop2006pattern} as the objective function as following.
\begin{IEEEeqnarray}{c} 
R(\hat{\pmb{T}})=\min_{\hat{\pmb{T}}_{1} \ge 0,...,\hat{\pmb{T}}_{I} \ge{0}}  \alpha N(\hat{\pmb{T}})+(1-\alpha) S(\hat{\pmb{T}}) \\
N(\hat{\pmb{T}})=\sum_{\tau=1}^{o}   \frac{1}{2}(\pmb{\epsilon}_{\tau}-\hat{\pmb{\epsilon}}_{\tau})^{\mathrm{T}}(\pmb{\epsilon}_{\tau}-\hat{\pmb{\epsilon}}_{\tau}) \IEEEnonumber\\
S(\hat{\pmb{T}})=KL(\hat{\pmb{d}}_{p}||\pmb{d}^{*}_{p})+KL(\hat{\pmb{d}}_{a}||\pmb{D}^{*}_{a}) \IEEEnonumber\\
where\  \hat{\pmb{\epsilon}}_{\tau}=   \sum_{k=1}^{t}  \pmb{P}_{\tau k} \hat{\pmb{T}}_{k}\IEEEnonumber
\end{IEEEeqnarray}

Our optimization process is shown in Algorithm 2. It is worth noting that since the optimization is alone the approximate distributions rather than the real distributions, the structure should not be optimal when $S(\hat{\pmb{T}})$ converges. Therefore, we then slack the structure constraint (set $\alpha$=1), which further does only numerical optimization and leads to a better point.

\begin{algorithm}  
  \caption{Optimization Algorithm}  
  \KwIn{ 
 Observed traffic counts tensor $\pmb{E}$\\
The best inferred OD distribution $\pmb{d}^{*}_{p}$ and $\pmb{d}^{*}_{a}$ from pre-trained Distribution Learner, separately.
  }  
  \KwOut{Estimated OD sequence
}
\textbf{Initialize}  Balance factor $\alpha$, Initialized OD sequence $\tilde{\pmb{T}}$, set $\hat{\pmb{T}}^{k}=\tilde{\pmb{T}}$ and $k=0^{th}$ iteration; \\
\Repeat{$R(\hat{\pmb{T}}^{k})$ convergent}  
{  
 \textbf{Lower level:} Simulate $\hat{\pmb{T}}^{k}$  with the simulator and obtain simulated trafﬁc counts observations $\pmb{\epsilon}_{\tau}, \tau=1,2,...o$. Assignment matrix $\pmb{P}$ is calculated by using a back-calculation procedure;\\ 
  \textbf{Upper level:} $\hat{\pmb{T}}^{k+1}=\min_{\hat{\pmb{T}}^{k}}R(\hat{\pmb{T}}^{k})$ based on Eq(14);\\
  \If{$S(\hat{\pmb{T}}^{k})$ convergent}{
  set $\alpha$=1
  }
  }
\end{algorithm}

\section{EXPERIMENTS AND RESULTS}
We test our method on a large-scale real city network with a synthetic dataset. The effectiveness of NN for distributional inference is first validated. Then, a comparative experiment is used to demonstrate the advantage of our optimization method compared with traditional numerical optimization. The project uses Python programming and relies on the TensorFlow system\cite{abadi2016tensorflow} for gradient calculation and NN modeling. Sklearn library\cite{pedregosa2011scikit} and Scipy\cite{virtanen2020scipy} are used for clustering and optimization program, respectively.

\subsection{Study Network}\label{AA}

We selected the $400km^{2}$ area around Cologne, Germany, as our study network and used SUMO as the simulator program to test on a large-scale city network\cite{uppoor2013generation}. 71368 road sections and 31584 intersections make up the network (Fig. 3(a)). We employ the K-means\cite{arthur2006k} technique to gather OD nodes on the network according to Euclidean distance. In this case, we select $n_{od}$=15 (Fig. 3(b)), and we aggregate the directed road sections from 782 to 64.

\begin{figure}
\centering
\subfigure[study network]{\includegraphics[width=4.3cm]{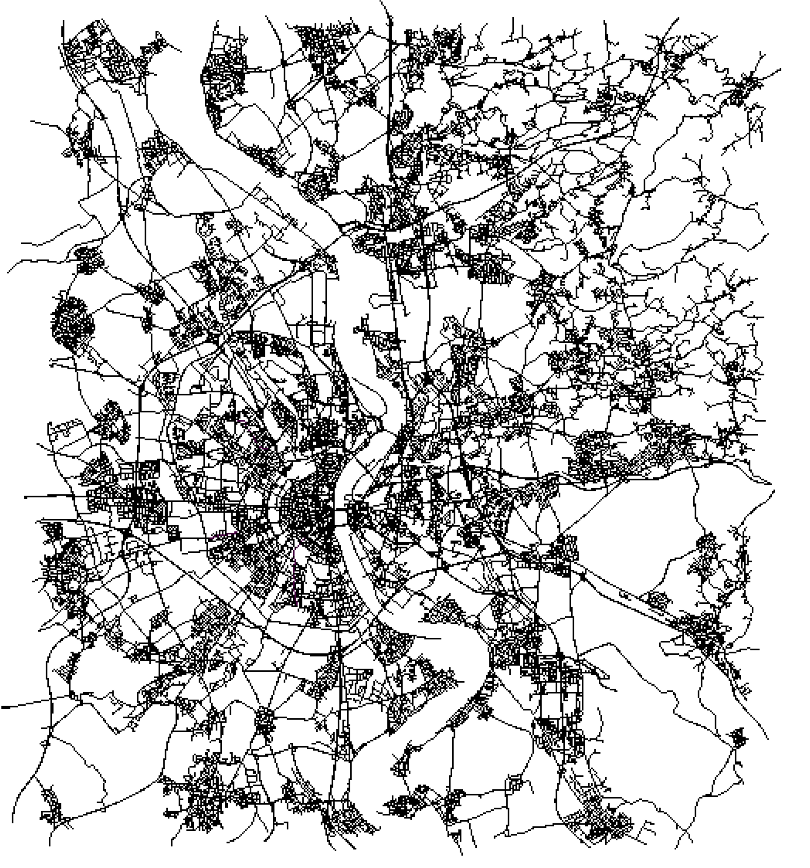}}
\subfigure[nodes and road sections]{\includegraphics[width=4.3cm]{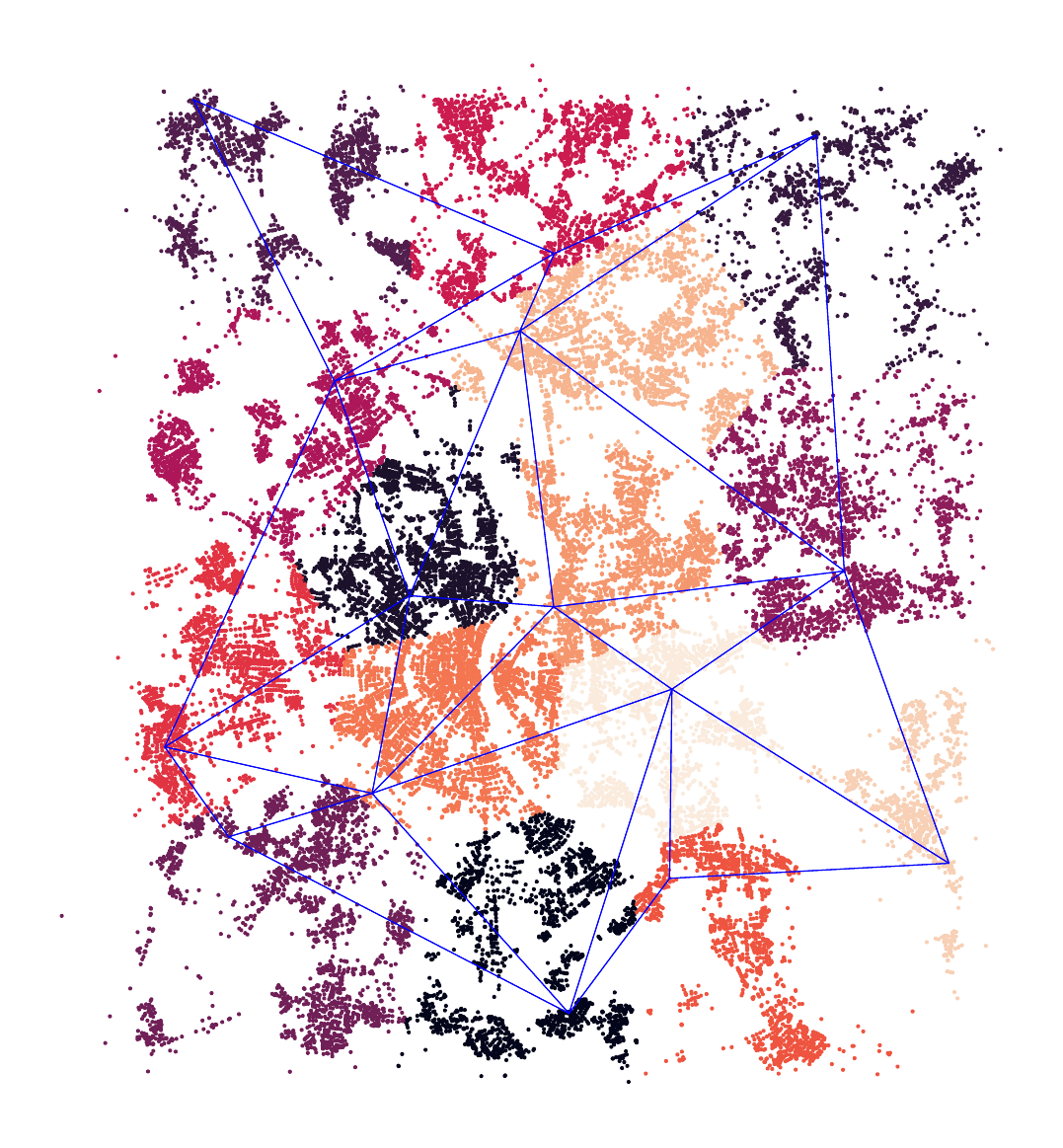}}
\caption{The aggregate traffic counts operation is denoising because vehicles will choose different road sections between two connected adjacent OD nodes.}
\label{fig:2 }  
\end{figure}

\begin{table*}
\centering
	\caption{ground truth OD sequence}
	\label{table}
	\setlength{\tabcolsep}{3pt} 
	\renewcommand\arraystretch{1.5} 
	\begin{tabular}{m{2.5cm}<{\centering}|m{1cm}<{\centering}|m{1cm}<{\centering}|m{1cm}<{\centering}|m{1cm}<{\centering}|m{1cm}<{\centering}|m{1cm}<{\centering}|m{1cm}<{\centering}|m{1cm}<{\centering}|m{1cm}<{\centering}|m{1cm}<{\centering}|m{1cm}<{\centering}|m{1cm}<{\centering}}
	\toprule

		OD matrix(o'clock)&$6-7$&$7-8$&$8-9$&$9-10$&$10-11$&$11-12$&$12-13$&$13-14$&$14-15$&$15-16$&$16-17$&$17-18$\\ \hline
		
		total travels & 76972 & 86580 & 40131 & 31971 & 34211 & 41065 & 50015 & 48899 & 48124 & 74760 & 95042 & 79205  \\ \hline
		density($>m$)  & 0.3689 & 0.3644 & 0.2844 & 0.2489 & 0.2533 & 0.2667 & 0.3111 & 0.3200 & 0.3022 & 0.3378 & 0.3644 & 0.3511  \\  \hline
		the largest OD pair & 9097 & 11048 & 4918 & 3395 & 3145 & 4523 & 5664 & 4643 & 4314 & 7937 & 10315 & 8816  \\  
		\bottomrule

	\end{tabular}
	\label{tab1}
\end{table*}

\subsection{Dataset}\label{AA}
The synthetic data closely comparable to the actual conditions of urban traffic serves as the ground truth. Refer to \cite{uppoor2013generation} for more information.

As shown in Table 2, we took the ground truth OD matrix for 12 hours from 6:00 am to 18:00 pm from a 24-hours traffic simulation. Set the OD estimation interval to 1 hour and the traffic counts observation interval to 10 minutes. For the input of the DCGRU module, we set $\delta$=4 to indicate that the traffic counts of the four flowing observation intervals from the current OD estimation interval are used as input to characterize the current OD matrix. Therefore, the whole length of observation intervals is 76 (12$\times$6 since each OD estimation interval is with six observation intervals)

We set $m$=50 and sample the OMS from the ground truth OD matrix. Then resample from the original matrices sequence to get various OD sequences, and generate the corresponding observation traffic counts $\pmb{\epsilon}$ to form a data set $(\pmb{D}_{E},\pmb{d}_{p})$ and $(\pmb{D}_{E},\pmb{d}_{a})$, the size of each is 10k sample pairs.

In an 8:2 ratio, we split the dataset for training and validation. A model is trained using the training data, and its generalization performance is assessed using the validated data.

\subsection{OD estimating evaluation}\label{AA}
Referring to\cite{behara2020novel}, we use the numerical value indicators  $RMSN(\hat{\pmb{T}}_{t},\pmb{T}_{t})$\cite{antoniou2004incorporating} and structural indicators $\rho(\hat{\pmb{T}}_{t},\pmb{T}_{t})$\cite{djukic2013reliability} to measure the gap between the estimated OD matrix $\hat{\pmb{T}}_{t}$ and the real OD matrix $\pmb{T}_{t}$.
 \begin{subequations}
\begin{equation}
RMSN(\hat{\pmb{T}}_{t},\pmb{T}_{t})=\frac{\sqrt{n_{od}^2\sum\limits^{n_{od}^2}_{i}(\pmb{T}_{ti}-\hat{\pmb{T}}_{ti})^2}}{\sum\limits^{n_{od}^2}_{i} \pmb{T}_{ti}}
\end{equation}
\begin{equation}
\rho(\hat{\pmb{T}}_{t},\pmb{T}_{t})=\frac{(\pmb{T}_{t}-\pmb{\mu})^{T}(\hat{\pmb{T}}_{t}-\pmb{\hat{\mu}})}{\sqrt{(\pmb{T}_{t}-\pmb{\mu})^{T}(\pmb{T}_{t}-\pmb{\mu})} \sqrt{(\hat{\pmb{T}}_{t}-\pmb{\hat{\mu}})^{T}(\hat{\pmb{T}}_{t}-\pmb{\hat{\mu}})} }
\end{equation}
 \end{subequations}
where $\pmb{\mu} \in \mathbb{R}^{n_{od}^{2}}_{\geq{0}}$ is a vector with each element value equal to the mean of $\pmb{T}_{t}$, and $\hat{\pmb{\mu}}$, $\tilde{\pmb{\mu}}$ corresponds to $\hat{\pmb{T}}_{t}$ and $\tilde{\pmb{T}}_{t}$, respectively.

\subsection{Parameter settings}\label{AA}
{
\begin{tabular}{ll}
  \hline
	head number $h$ & 6 \\
	encoder layer $N$ & 2 \\
	learning rate $r$ & 1E-4 \\
	dimention $d$ & 128\\
	diffusion convolution step $K$ & 2\\
	maximum trips value $m$ & 50\\
	OD estimation intervals  $I$ & 12\\
	observation intervals $o$ & 12 and 72\\
	sequence length $\delta$ & 4\\
	\hline
\end{tabular}
}

\subsection{Results and analyze}\label{AA}
\subsubsection{Training}

Firstly,  As shown in Fig. 4. We set the real OD sequence as our test set, the test set curve converges indicates our sampled training data is effective for model training. Moreover, the results of the test set is not as good as training and validation set since the distribution of real data and sampled data is not perfectly consistent. Which is a common problem in deep learning and implies it can be further alleviated through more appropriate sampling methods.

Secondly, we tested the impact of different observation intervals length $\delta$ on the inferred results. Our experiment in Fig. 5 shows that, when $\delta=4$ the inferenced results by NN model is the best. It means our model does not completely utilize all the observation intervals information during one estimation interval(should be $\delta=6$), and when there is an overlap ($\delta>6$), the results get worse. It indicates that our current model has not effectively extract all the information introduced by longer observation intervals, more advanced NN model could further improve the inference results.
\begin{figure}
\centering
\subfigure[$Loss_{p}$]{\includegraphics[width=4.3cm]{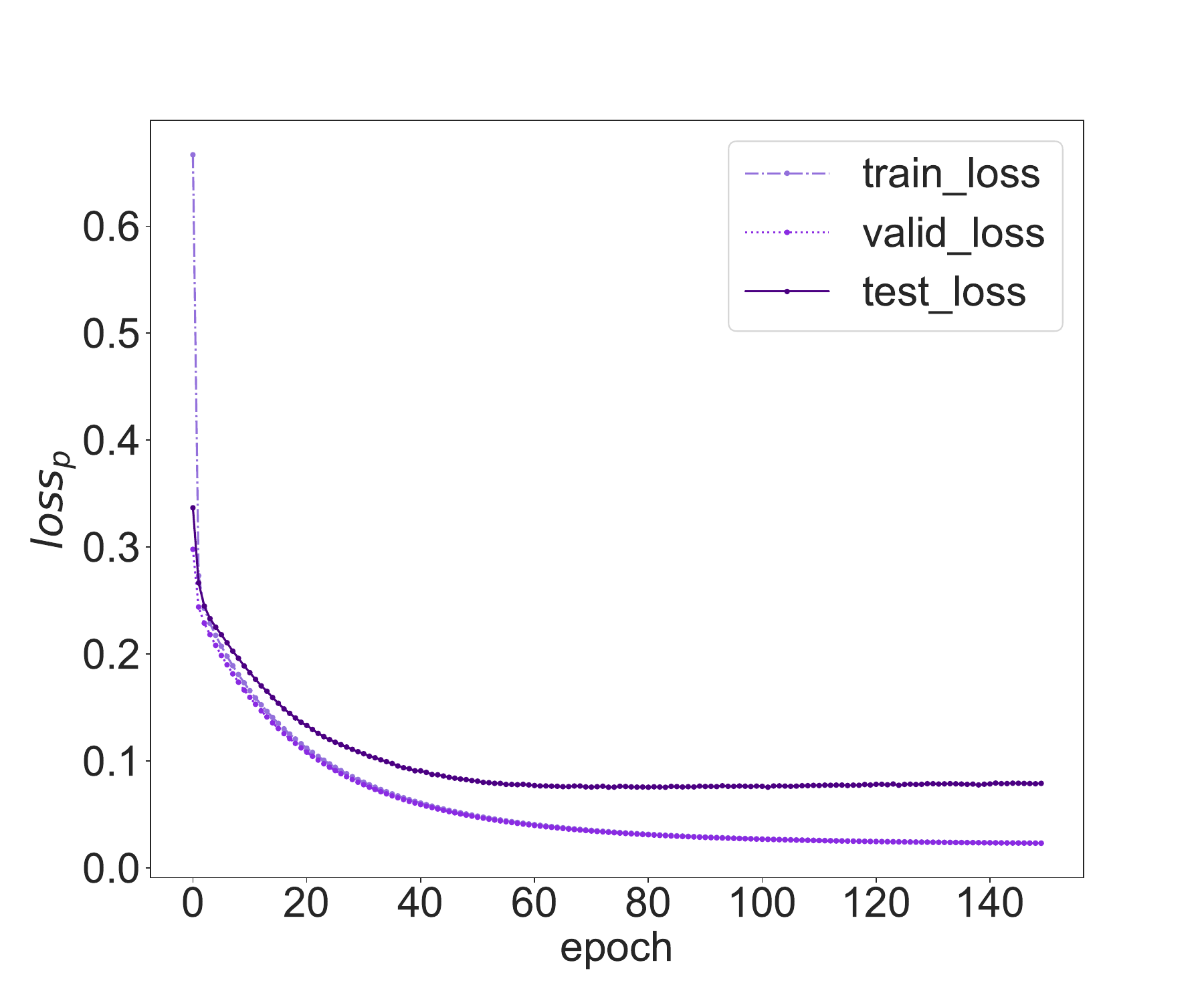}}
\subfigure[$Loss_{a}$]{\includegraphics[width=4.3cm]{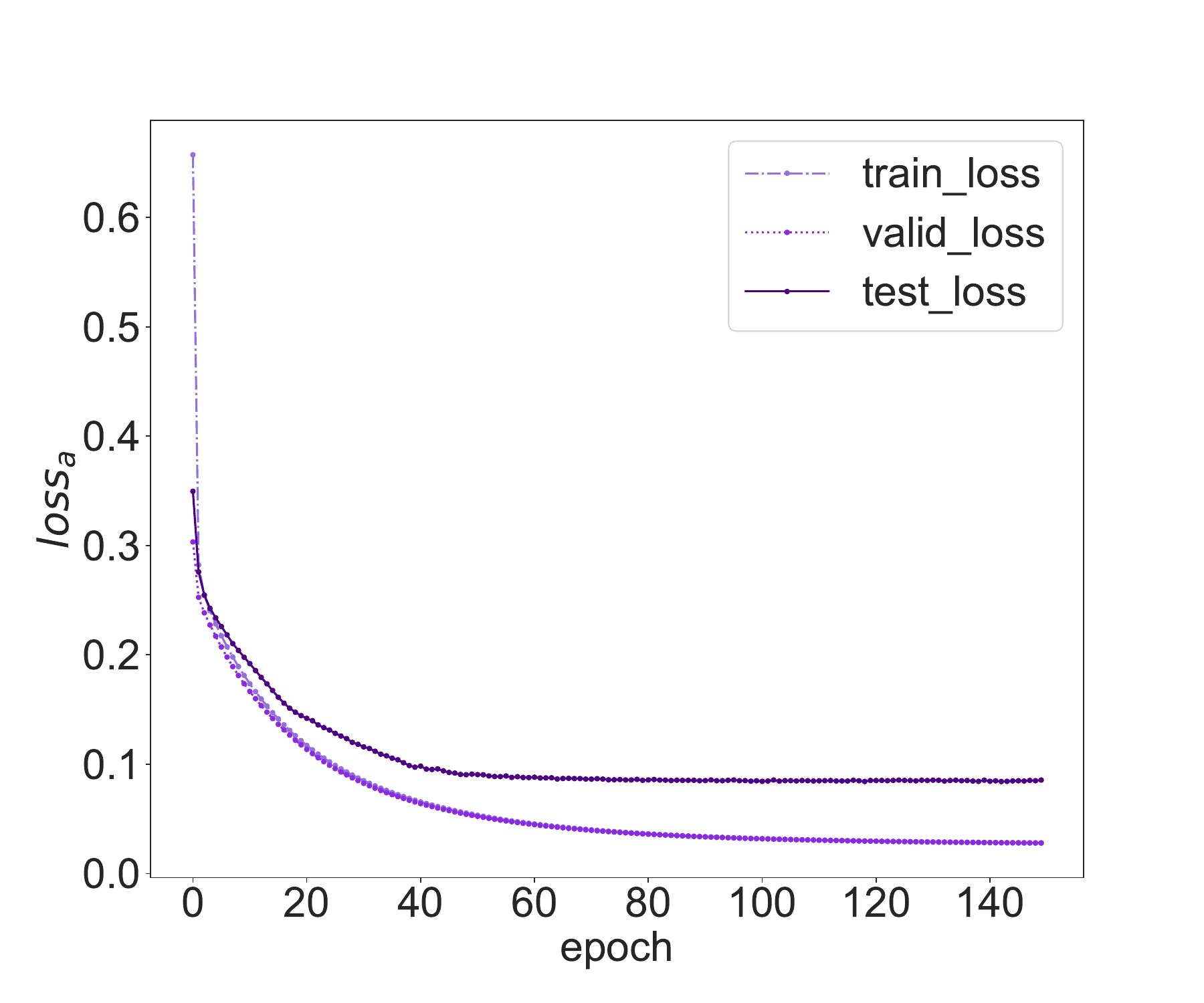}}
\caption{training loss curves for the Distribution Learner} 
\label{ }  
\end{figure}

\begin{figure}
\centering
{\includegraphics[width=8cm]{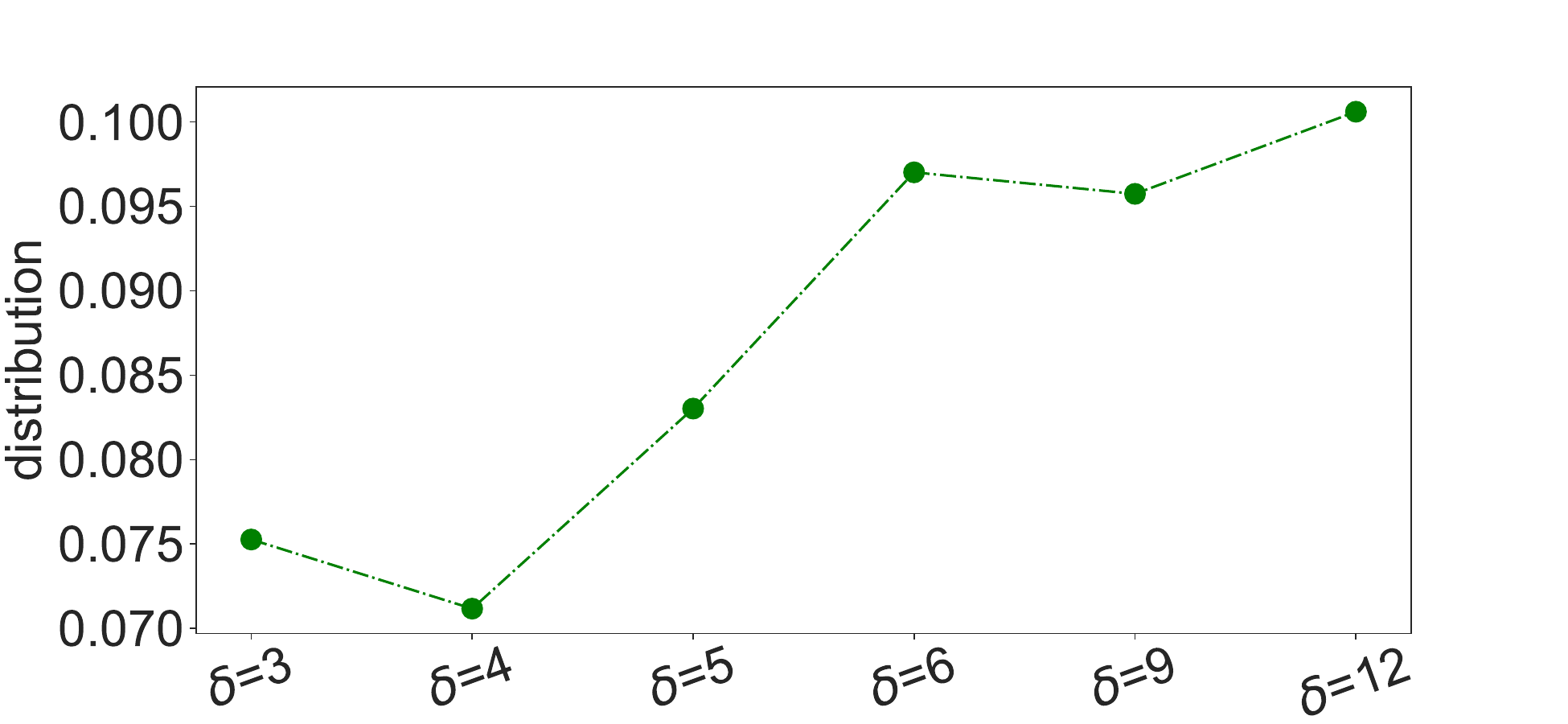}}
\caption{the impact of different observation sequences length $\delta$ on the inferred results} 
\label{} =
\end{figure}

\begin{figure*}
\centering
\subfigure[production distribution expanding on the time series]{\includegraphics[width=8cm]{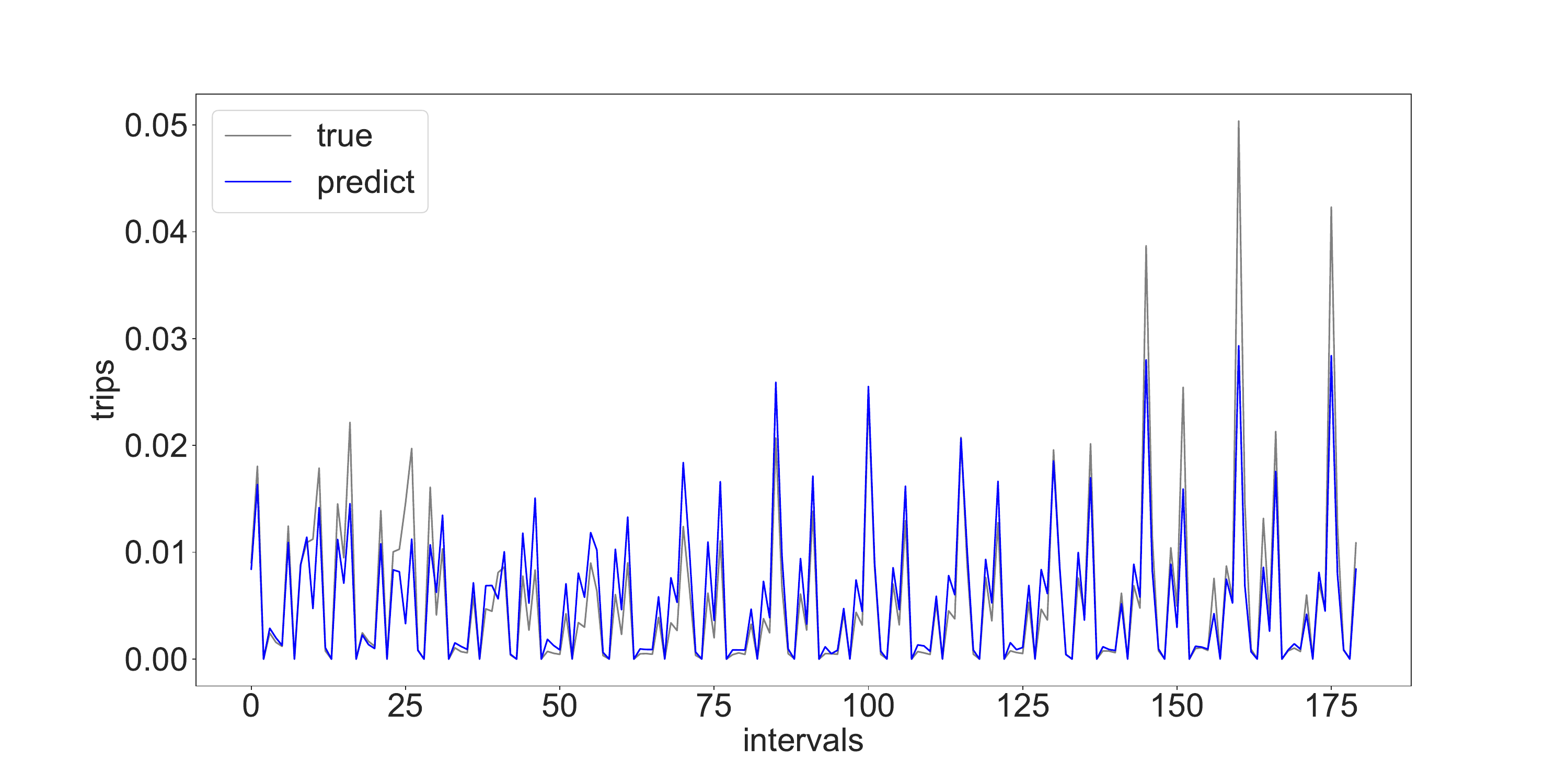}}
\subfigure[spatio distribution of OD nodes when 12-13 o'clock]{\includegraphics[width=8cm]{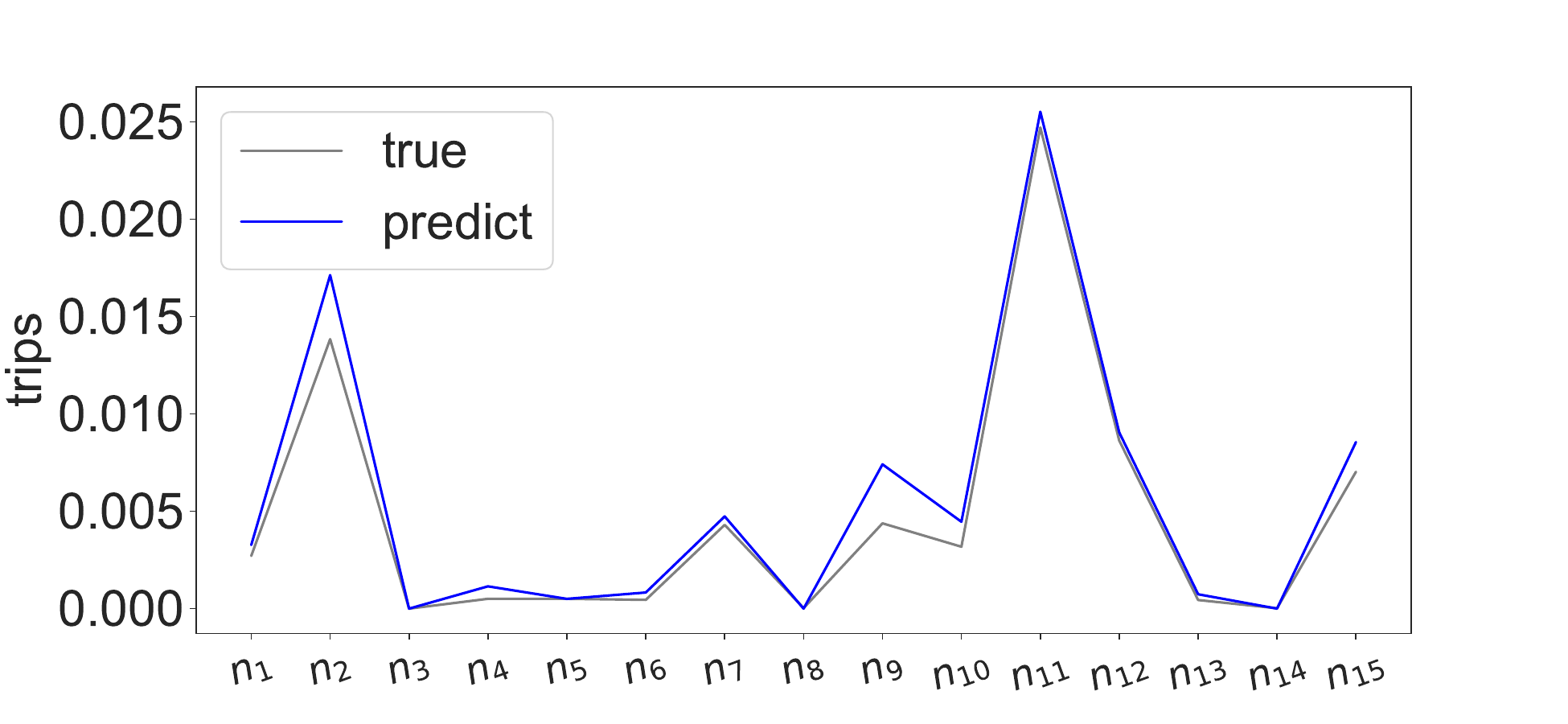}}
\caption{production distribution inferred} 
\label{}  
\end{figure*}

\begin{figure*}
\centering
\subfigure[1e-2 ]{\includegraphics[width=4.4cm]{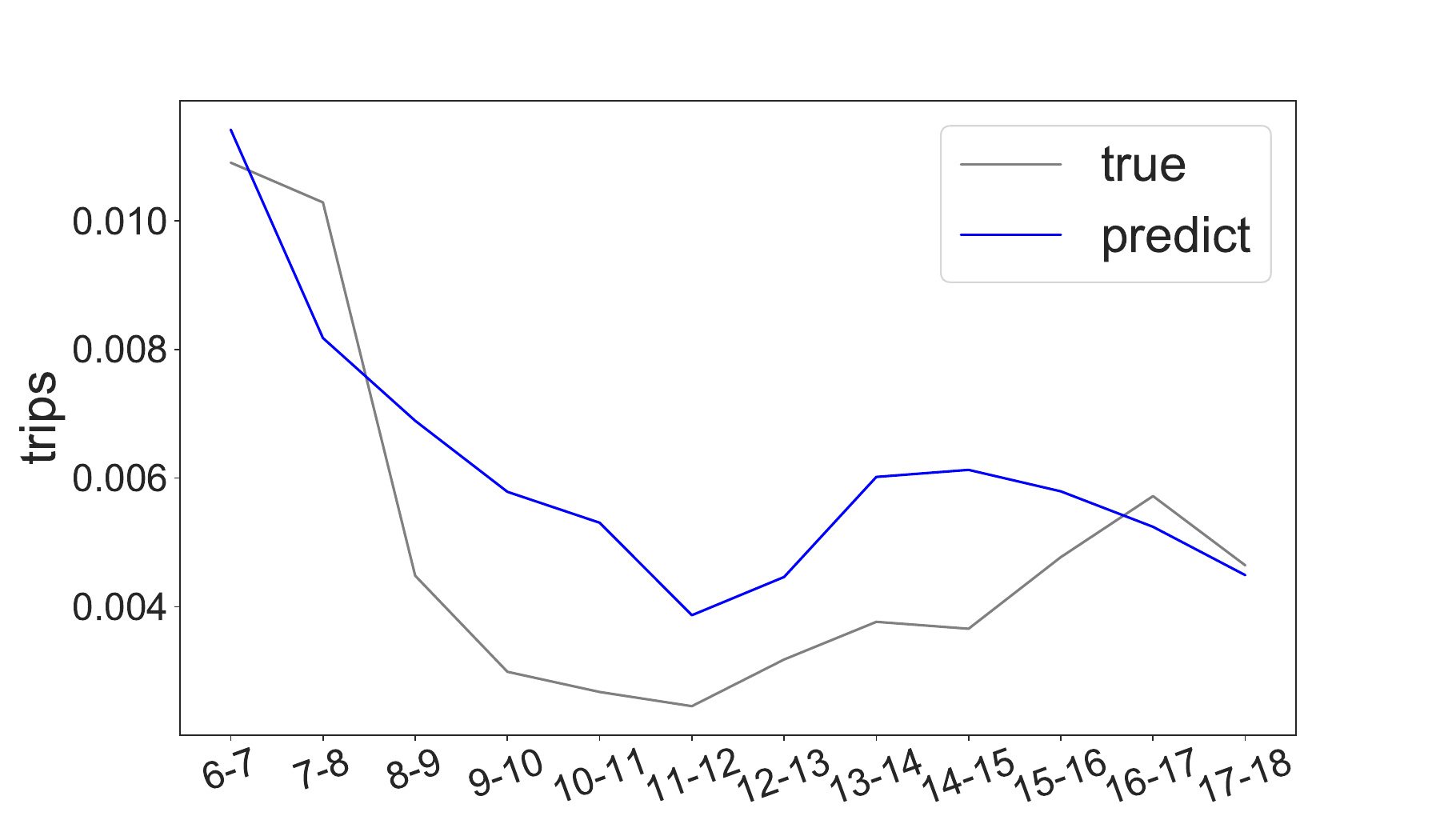}}
\subfigure[1e-2]{\includegraphics[width=4.4cm]{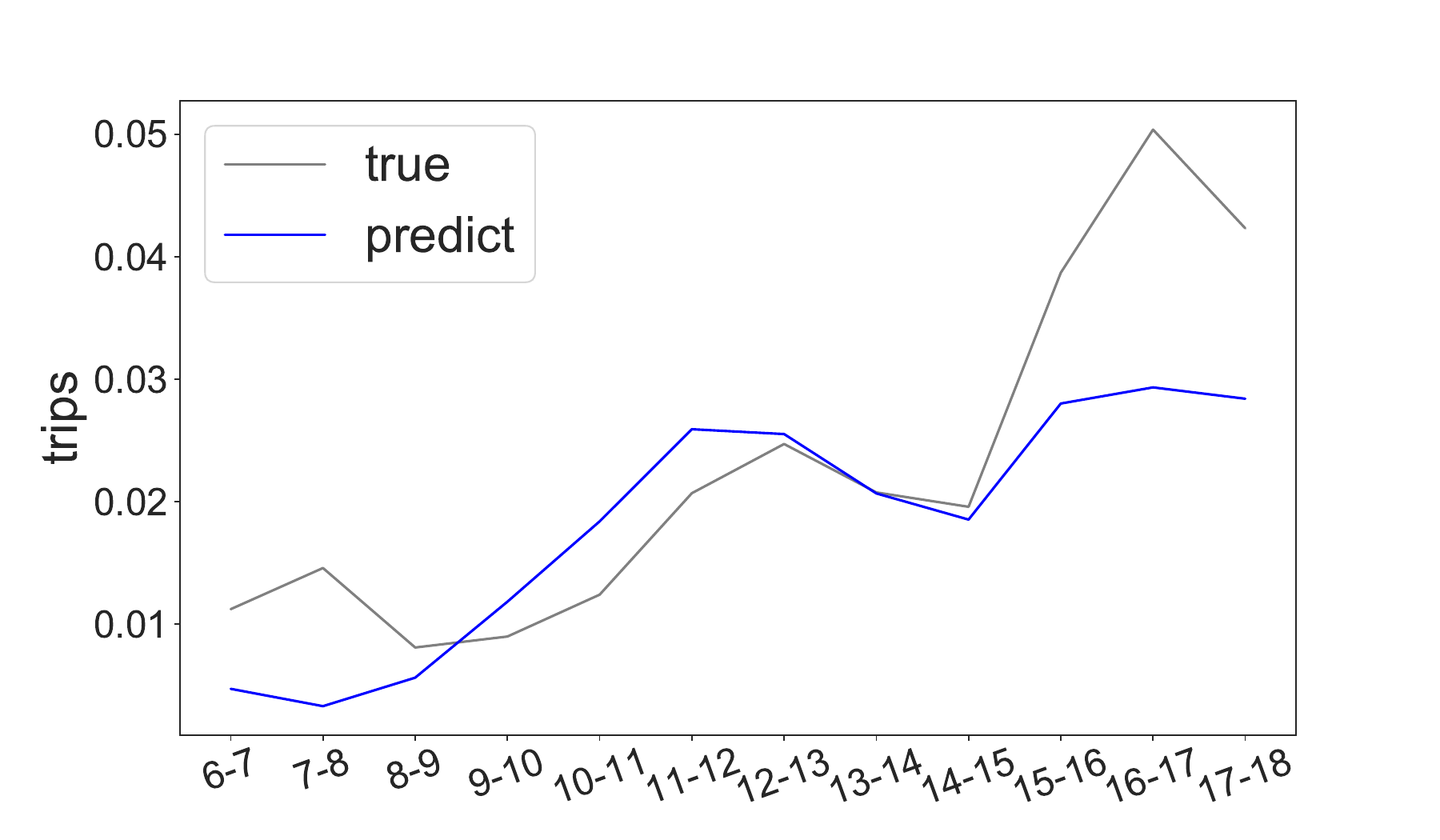}}
\subfigure[1e-3]{\includegraphics[width=4.4cm]{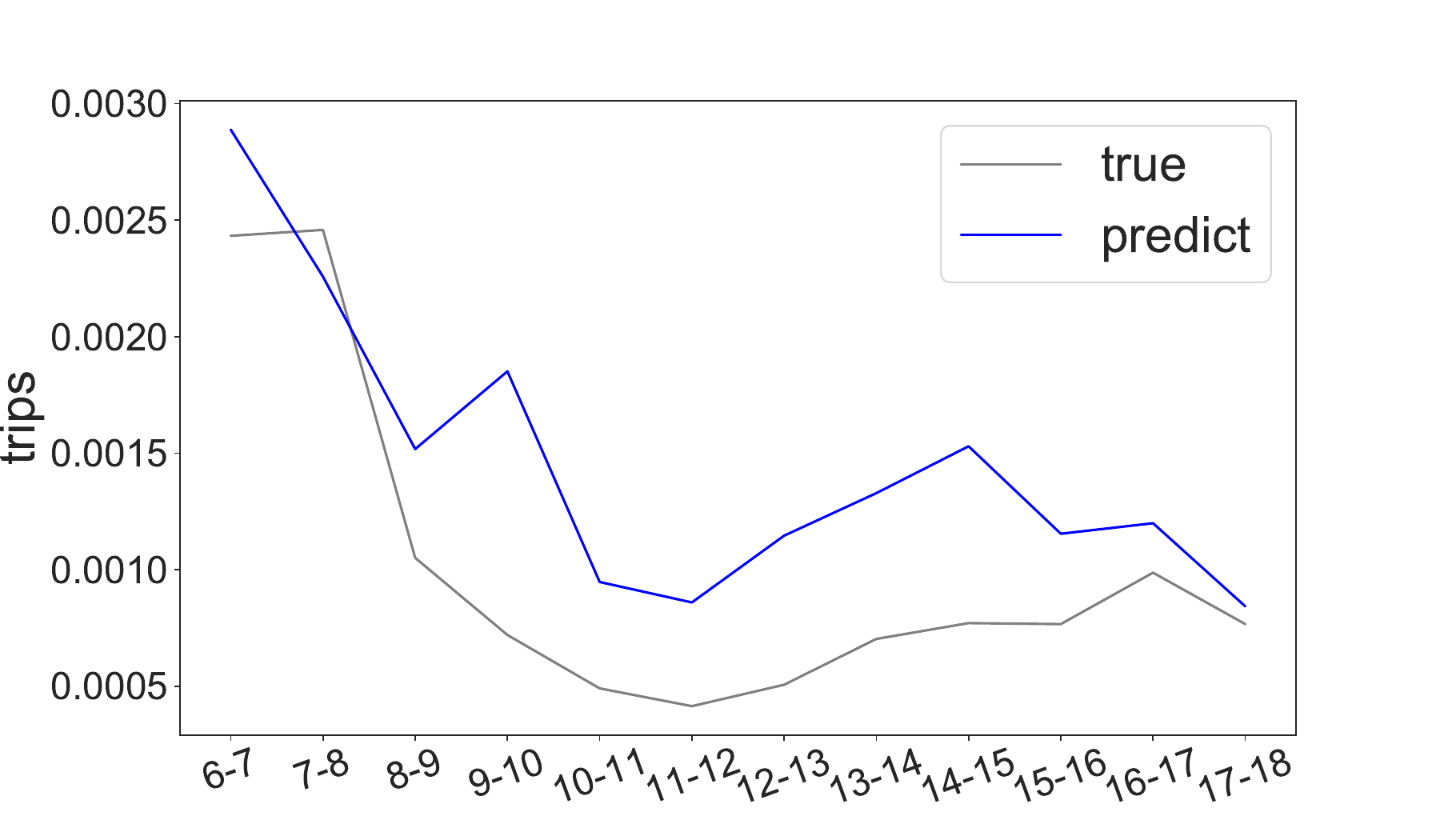}}
\subfigure[1e-5]{\includegraphics[width=4.4cm]{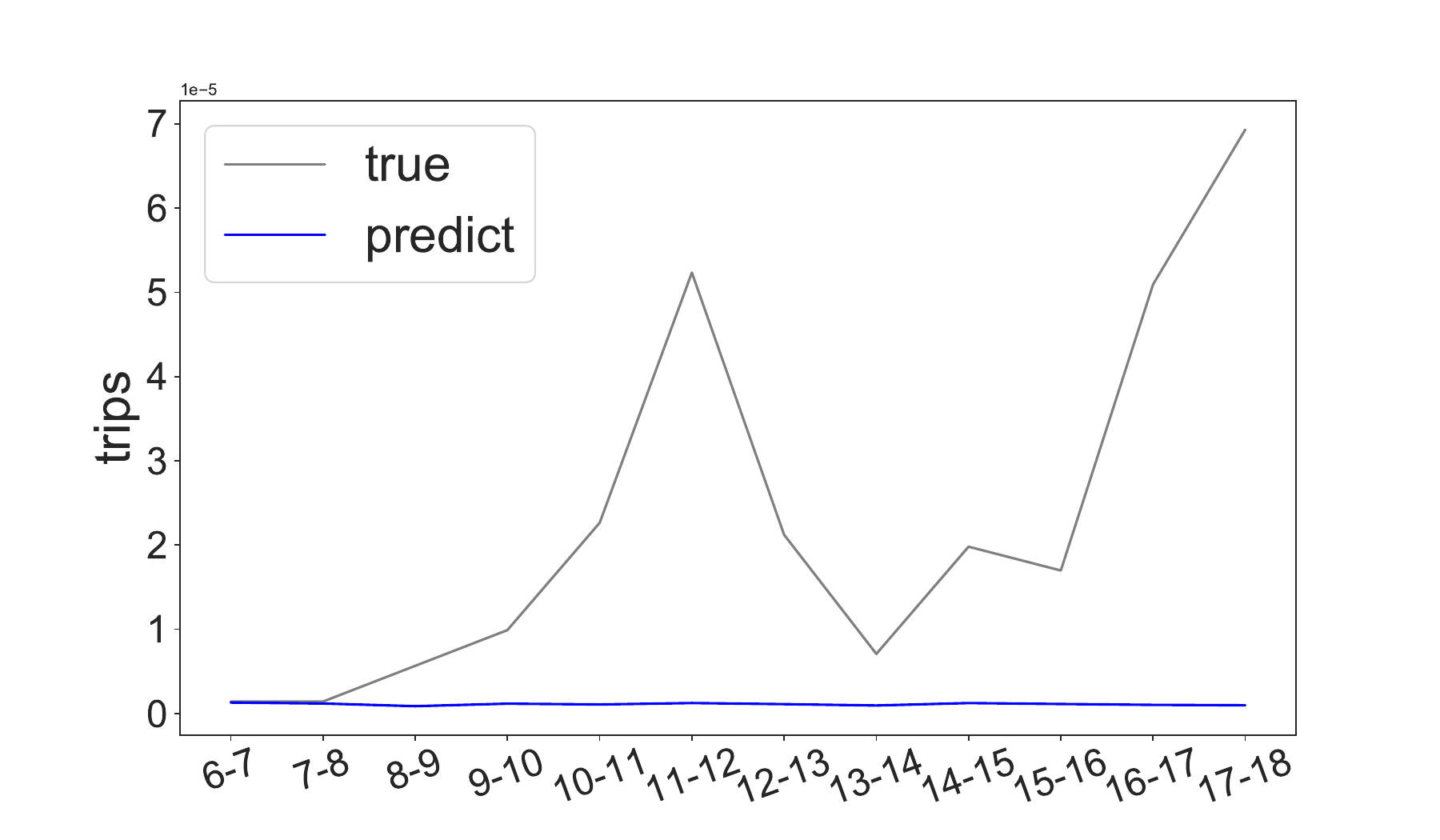}}

\caption{temporal evolution of specific OD node (a)Node9 (b)Node10 (c)Node3 (d)Node13} 
\label{}  
\end{figure*}


\subsubsection{Distribution Inference}
Firstly, Fig. 6(a) shows the best inference results of our NN model on the global production distribution $\pmb{d}_{p}^{*}$, it reflects the ability of our model to give the global spatio-temporal distribution of OD sequence. Secondly, we pick an OD matrix at 12-13 o’clock from the global distriution, as shown in Fig. 6(b), which shows that the model has a good performance on spatial distribution inference, it clearly reflects the proportional between different OD nodes in the same OD matrix. Finally, we picked the distributions of four OD nodes during the whole estimation period, as shown in Fig. 7, the model infers the temporal evolution trend of specific OD node also very well, moreover in Fig. 7 we demonstrate with different size of proportions(from le-2 to le-5) and it shows our model has a high sensitivity that, except 1e-5, since these proportions are too small to cause the model to infer the node has a proportions of 0.

\subsubsection{Optimization}
Firstly, we compare two traditional optimization methods. These methods only use numerical optimization with two different observation interval sets: one hour and 10 minutes, respectively. The one hour setting indicates there are 12 observation intervals since we have 12 hours during the whole estimation period, so we refer to this method as Traditional($o$=12). And another is Traditional($o$=72) since there are 72 observation intervals for 10 minutes setting. Although 10 minutes setting has the smallest scale and can provides more constraints to alleviate the underdetermined problem, as mentioned in\cite{cascetta1993dynamic}, it also introduces more noise in the optimization process, which may leading to a poor optimization result. As seen in Table 3, the final optimization result obtained by the 10 minutes setting is worse than the one hour setting, which indicates that the noise in the optimization process offset the benefits of excessive observation intervals. Secondly, our method namd Ours with $\pmb{d}^{*}$ can mining significant information from these small sacle observation data to infer accurate global distributions $\pmb{d}_{p}^{*}$ and $\pmb{d}_{a}^{*}$ and then guide optimization process to find a better result. In addition, we also provide the results obtained from Ours with $\pmb{d}$, which utilizing the real global distribution $\pmb{d}_{p}$ and $\pmb{d}_{a}$ as guide, and provides the upper bound of our method, which is shown in Fig. 8. It can indicate that our method has a potent extendibility, such as through better data sampling methods or better models to infer more accurate distributions and get better optimization results.

\begin{figure*}
\centering
\subfigure[curve of $\rho$]{\includegraphics[width=4.4cm]{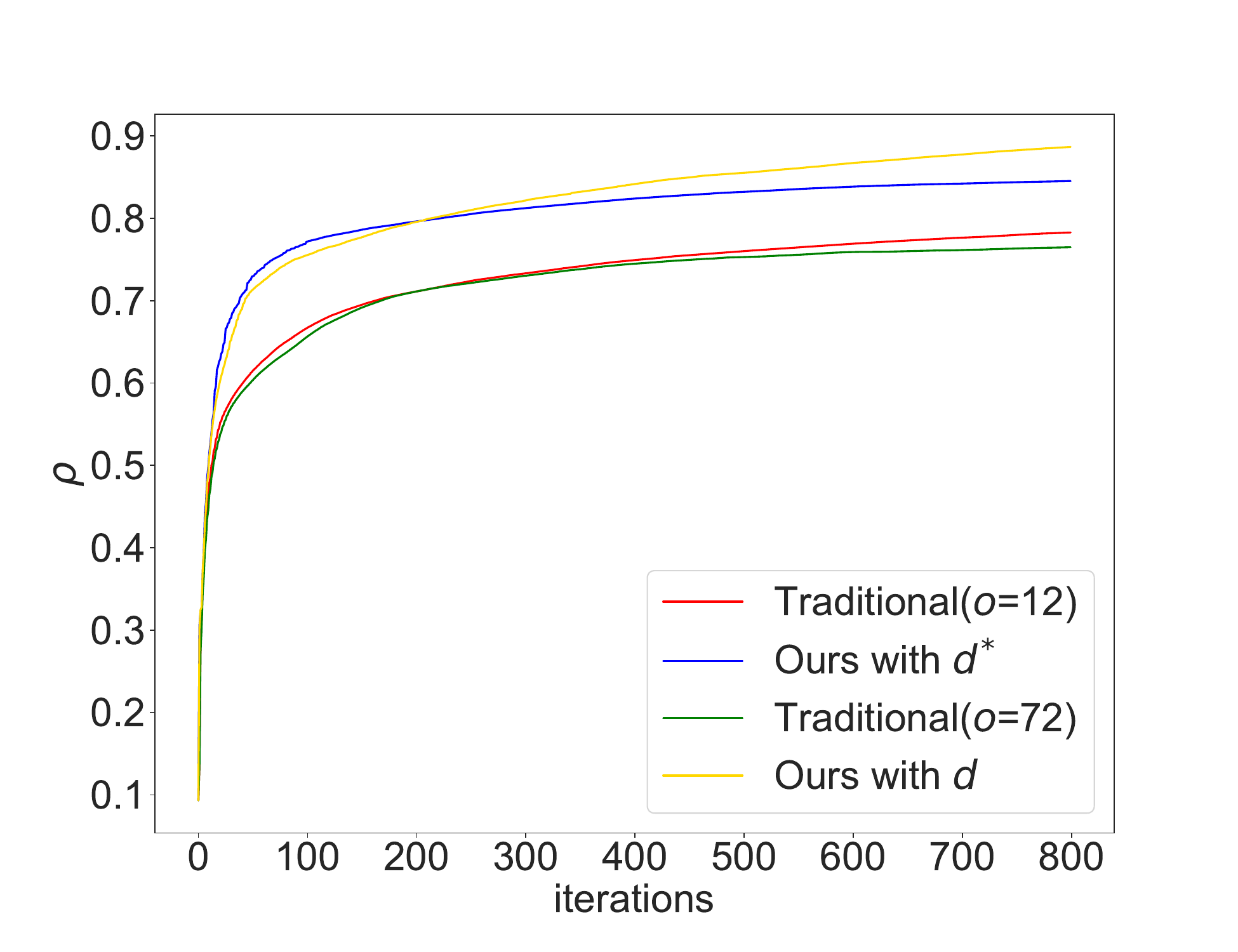}}
\subfigure[curve of $RMSN$]{\includegraphics[width=4.4cm]{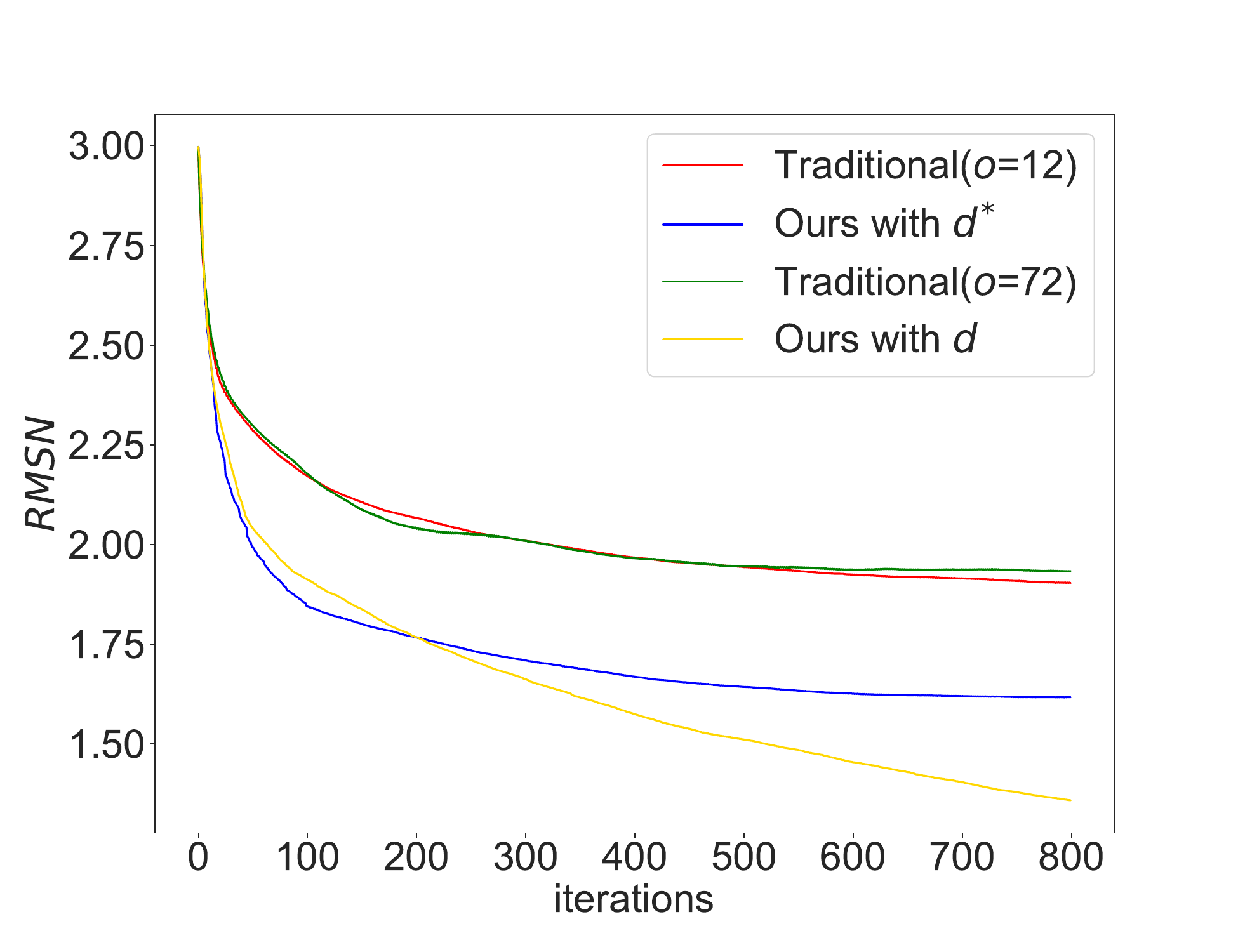}}
\subfigure[curve of $KL(\hat{d}_{p}||d_{p})$]{\includegraphics[width=4.4cm]{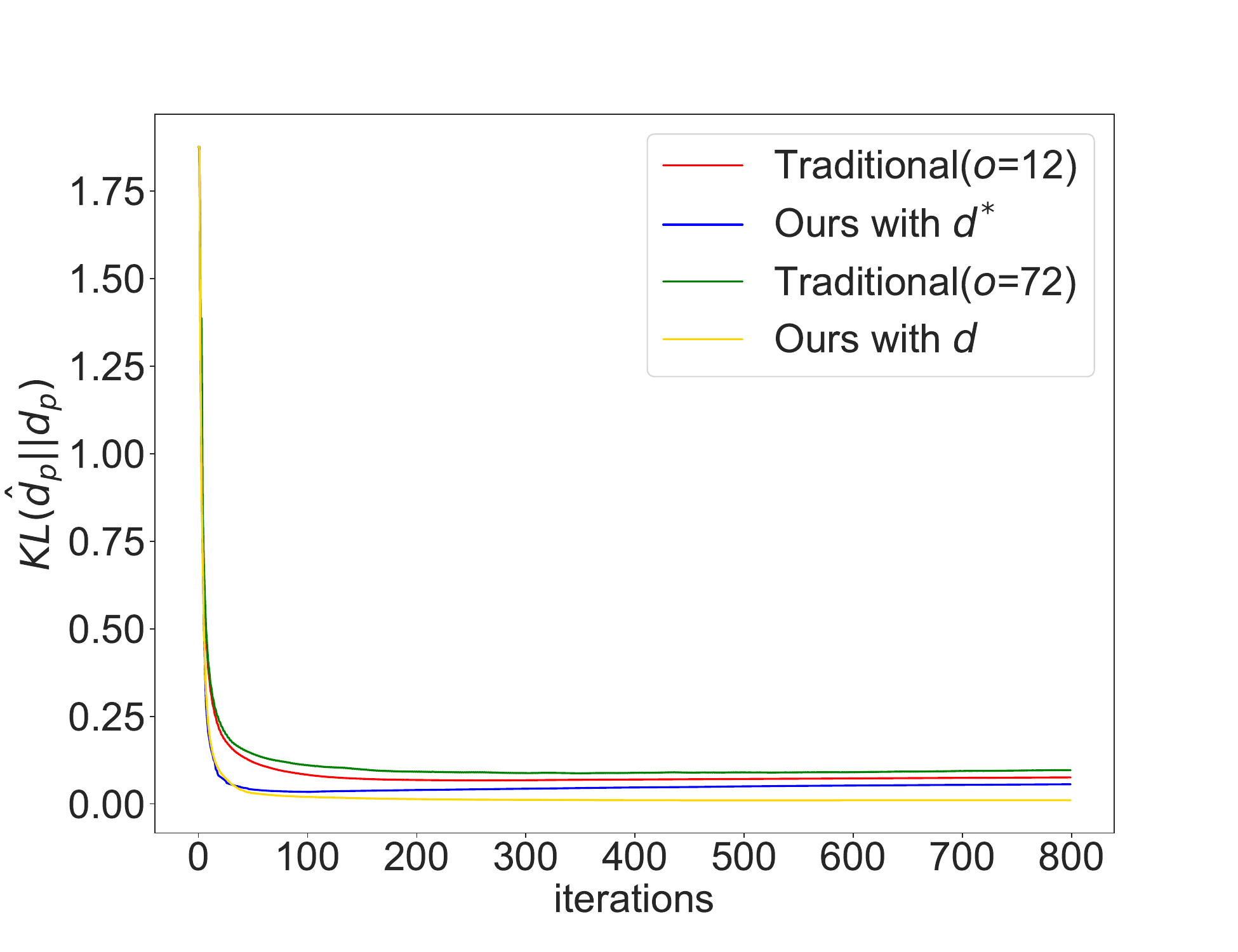}}
\subfigure[curve of $KL(\hat{d}_{a}||d_{a})$]{\includegraphics[width=4.4cm]{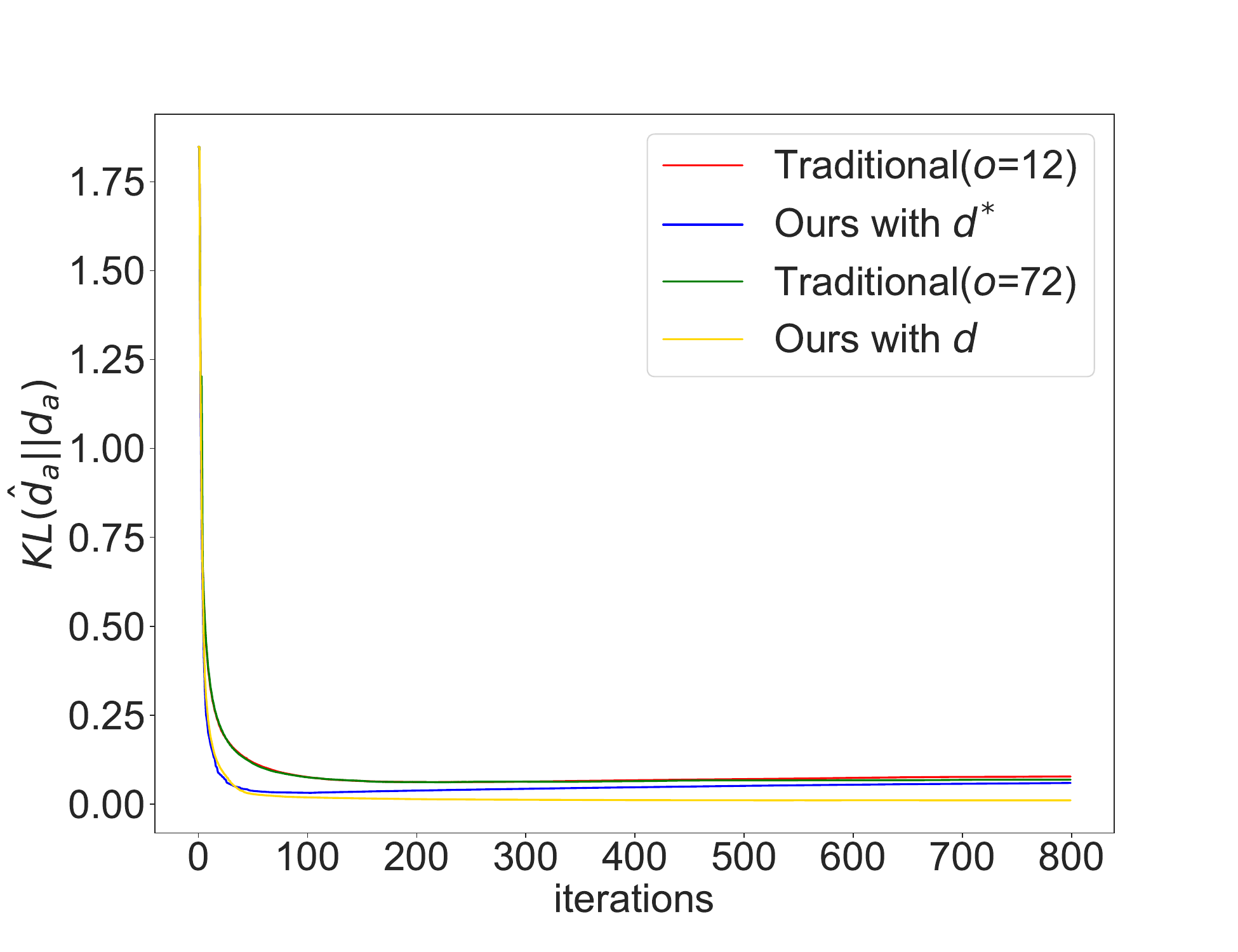}}
\caption{metrics for the optimization} 
\label{} 
\end{figure*}

\begin{table*}
\centering
	\caption{Optimization result}
	\label{table}
	\setlength{\tabcolsep}{3pt} 
	\renewcommand\arraystretch{1.5} 
	\begin{tabular}{m{2cm}<{\centering}|m{1cm}<{\centering}|m{1cm}<{\centering}|m{1cm}<{\centering}|m{1cm}<{\centering}|m{1cm}<{\centering}|m{1cm}<{\centering}|m{1cm}<{\centering}|m{1cm}<{\centering}|m{1cm}<{\centering}|m{1cm}<{\centering}|m{1cm}<{\centering}|m{1cm}<{\centering}|m{1cm}<{\centering}}
	\toprule

		OD matrix(o'clock)&$6-7$&$7-8$&$8-9$&$9-10$&$10-11$&$11-12$&$12-13$&$13-14$&$14-15$&$15-16$&$16-17$&$17-18$&Average\\ \hline

		  \multicolumn{14}{c}{$RMSN$} \\ \hline  
		Traditional($o$=12) & 1.77 & 1.88 & 1.76 & 1.81 & 1.96 & 2.02 & 2.00 & 1.90 & 1.82 & 2.11 & 2.17 & 1.60 & 1.90  \\ 
		Ours with $\pmb{d}^{*}$ & 1.68 & 1.92 & 1.27 & 1.21 & 1.32 & 1.55 & 1.60 & 1.46 & 1.43 & 2.04 & 2.30 & 1.57 & 1.61 \\ 
		Traditional($o$=72) & 2.09 & 2.06 & 2.49 & 1.62 & 1.84 & 2.20 & 1.81 & 1.25 & 2.52 & 2.39 & 1.40 & 1.45 & 1.93 \\ 
		Ours with $\pmb{d}$ & 1.33 & 1.51 & 1.01 & 1.51 & 1.43 & 1.41 & 1.41 & 1.35 & 1.15 & 1.44 & 1.51 & 1.18 & 1.35 \\  \hline
		 
		  \multicolumn{14}{c}{$\rho$} \\ \hline  
		Traditional($o$=12) & 0.8470 & 0.8358 & 0.8149 & 0.7729 & 0.7266 & 0.7502 & 0.7681 & 0.7539 & 0.7573 & 0.7465 & 0.7615 & 0.8583 & 0.7828 \\ 
		Ours with $\pmb{d}^{*}$  & 0.8567 & 0.8167 & 0.9096 & 0.9066 & 0.8851 & 0.8614 & 0.8517 & 0.8585 & 0.8560 & 0.7618 & 0.7182 & 0.8610 & 0.8453\\ 
		Traditional($o$=72) & 0.7496 & 0.8677 & 0.5921 & 0.8245 & 0.7661 & 0.6735 & 0.8020 & 0.9255 & 0.5535 & 0.6182 & 0.9055 & 0.8999 & 0.7649 \\ 
		Ours with $\pmb{d}$ & 0.9067 & 0.8824 & 0.9435 & 0.8477 & 0.8579 & 0.8764 & 0.8818 & 0.8763 & 0.9064 & 0.8745 & 0.8685 & 0.9188 & 0.8867 \\  \hline
		\bottomrule

	\end{tabular}
	\label{tab1}
\end{table*}

\begin{table*}
\centering
	\caption{KLD results in inference and optimization phases}
	\label{table}
	\setlength{\tabcolsep}{3pt}
	\renewcommand\arraystretch{1.5} 
	\begin{tabular}{m{2cm}<{\centering}|m{2.5cm}<{\centering}||m{2.5cm}<{\centering}|m{2.5cm}<{\centering}|m{2.5cm}<{\centering}|m{2.5cm}<{\centering}|m{2.5cm}<{\centering}}
	\toprule
	
		 Inference& Inferred &Optimization&Traditional($o$=12)&Ours with $\pmb{d}^{*}$&Traditional($o$=72)&Ours with $\pmb{d}$\\ \hline
		
		$KL(\pmb{d}_{p}^{*}||\pmb{d}_{p})$   & 0.0711 & $KL(\hat{\pmb{d}}_{p}||\pmb{d}_{p})$ & 0.0756 & 0.0560 & 0.0966 & 0.0105  \\ \hline
		$KL(\pmb{d}_{a}^{*}||\pmb{d}_{a})$   & 0.0826 & $KL(\hat{\pmb{d}}_{a}||\pmb{d}_{a})$ & 0.0772 & 0.0591 & 0.0684 & 0.0102 \\  
		\bottomrule

	\end{tabular}
	\label{tab1}
\end{table*}

In Fig. 9, we demonstrate the estimation results of four OD pairs along time series with four different orders of magnitude(1e1 to 1e4), respectively, to illustrate that our method has better performance on OD squence estimation tasks of different orders of magnitude. In there we only choose Traditional($o$=12) and Ours with $\pmb{d}^{*}$ since Traditional($o$=12) has the better performance in numerical methods, and Ours with $\pmb{d}^{*}$ is an only prctical way to use approximate distributions $\pmb{d}^{*}_{p}$ and $\pmb{d}^{*}_{a}$ from our distribution learner. And in Fig. 10 we show x-y plots of the estimation results for four different OD matrices.

As shown in Table 4, it is worth noting that the KLD of the best distribution $\pmb{d}_{p}^{*}$ inferred by the our NN model is 0.0711. Our actual results from the final optimization is $KL(\hat{\pmb{d}}_{p}||\pmb{d}_{p})$=0.0560. And KLD of the best distribution $\pmb{d}_{p}^{*}$ inferred by the our NN model is 0.0826, and our actual results from the final optimization is $KL(\hat{\pmb{d}}_{a}||\pmb{d}_{a})$=0.0591, which are much smaller than the approximate distributions be given since the approximate distributions are only used as a guide in the optimization process. Moreover, our convergence speed is also faster than traditional methods. It indicates that using the approximate distribution as a guide can help the optimization converge to a better point and faster.

\begin{figure*}
\centering
\subfigure[1e4]{\includegraphics[width=5cm]{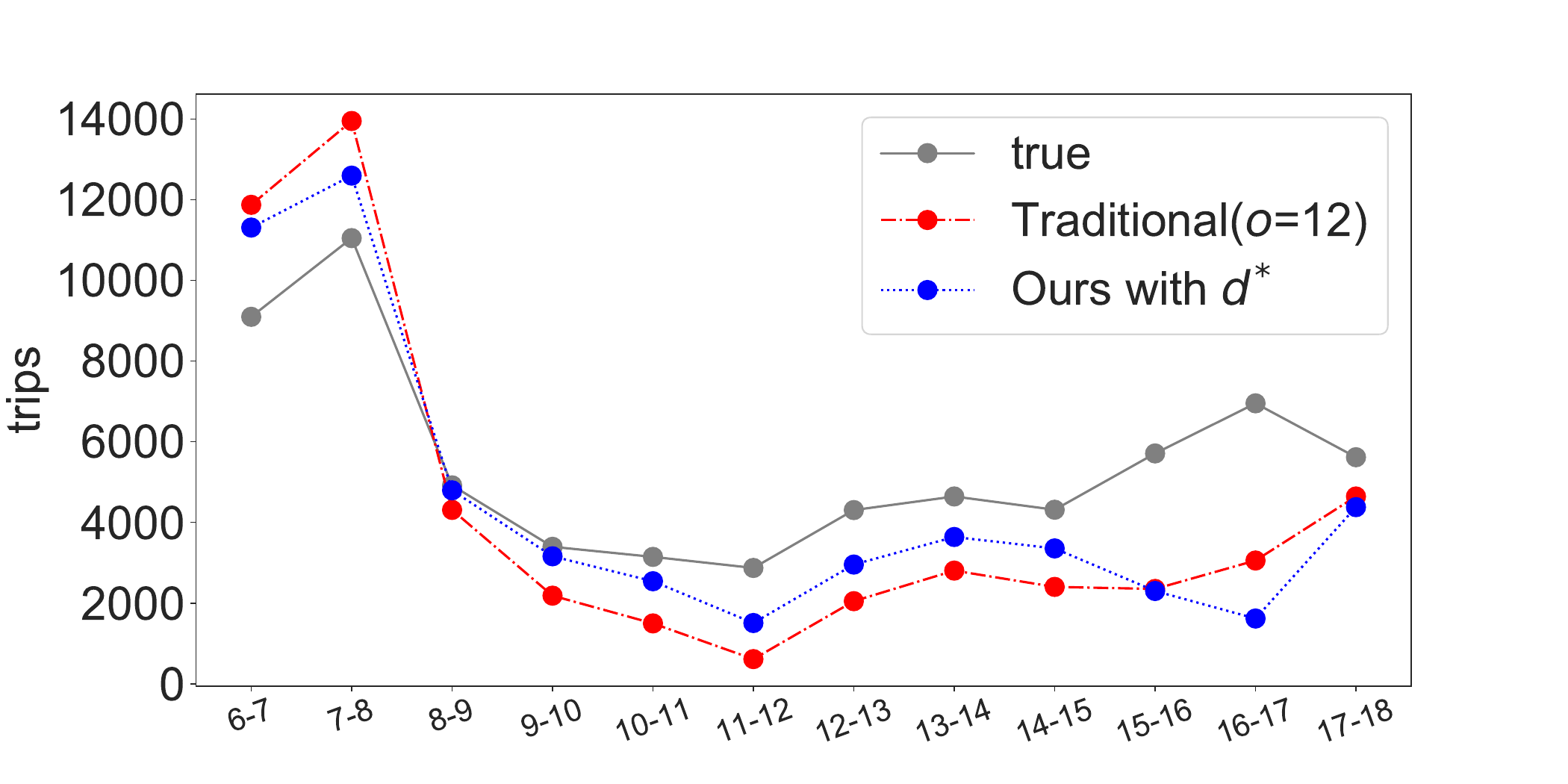}}
\subfigure[1e3]{\includegraphics[width=5cm]{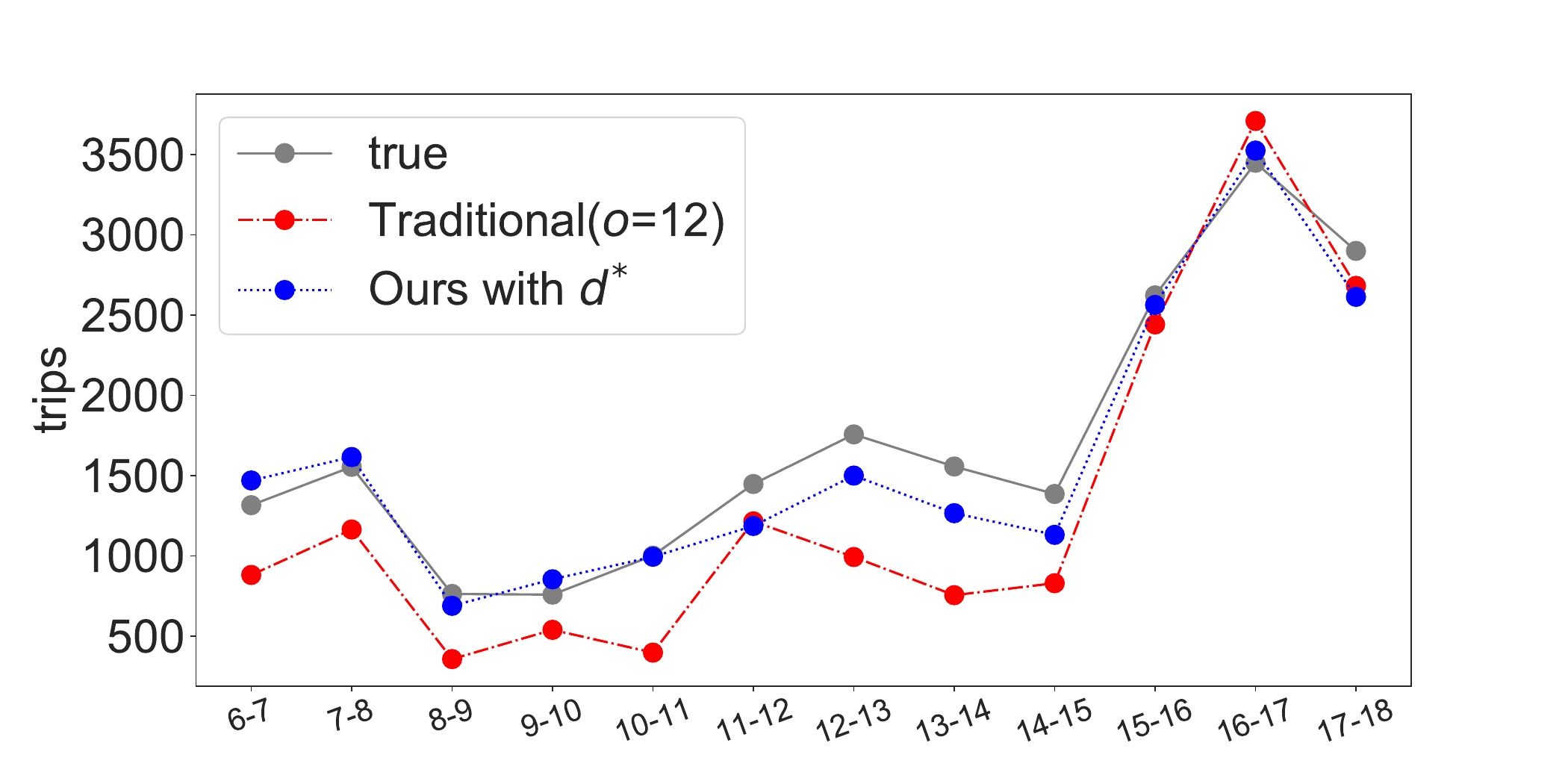}}
\subfigure[1e2]{\includegraphics[width=5cm]{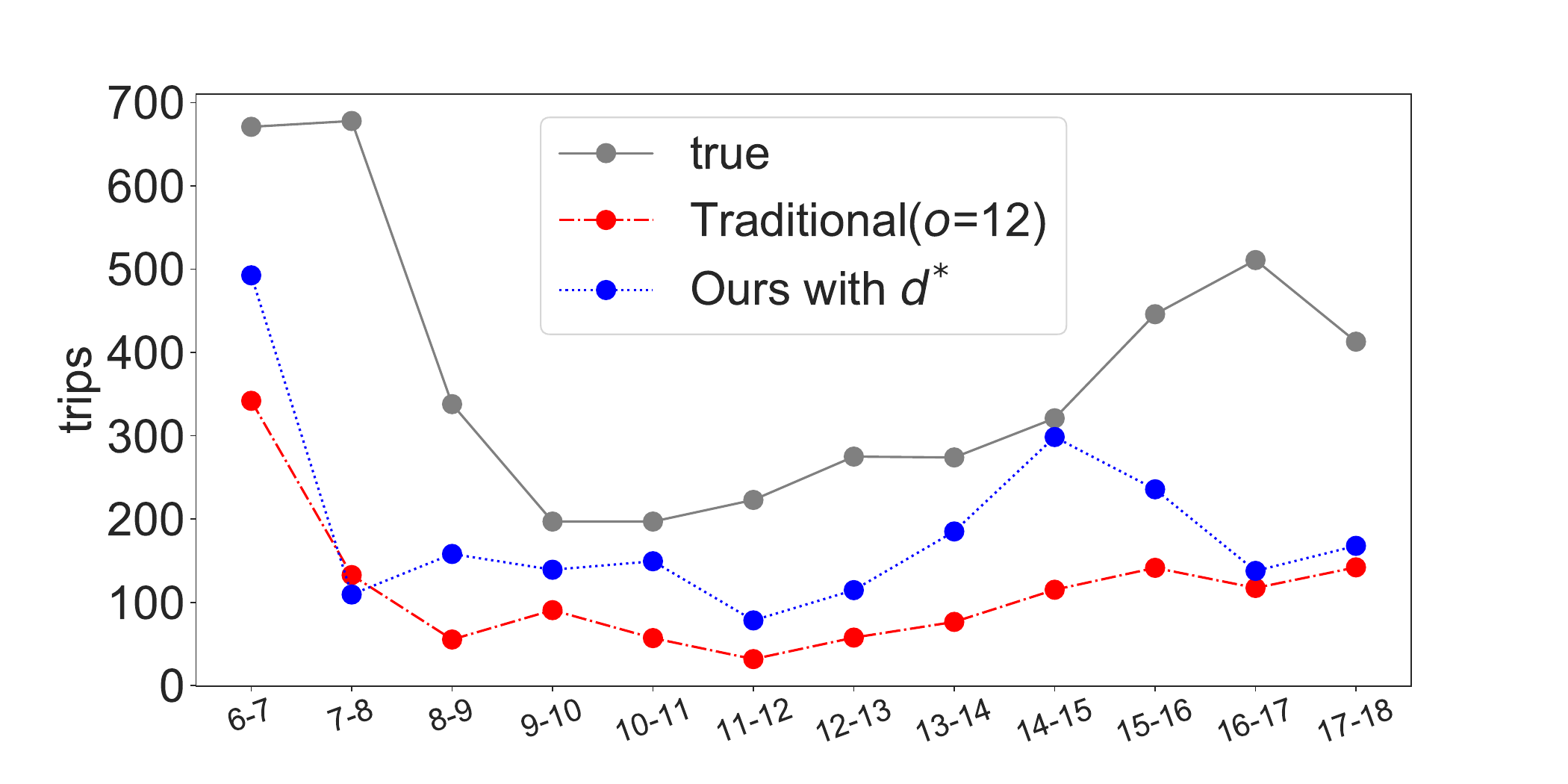}}
\subfigure[1e1]{\includegraphics[width=5cm]{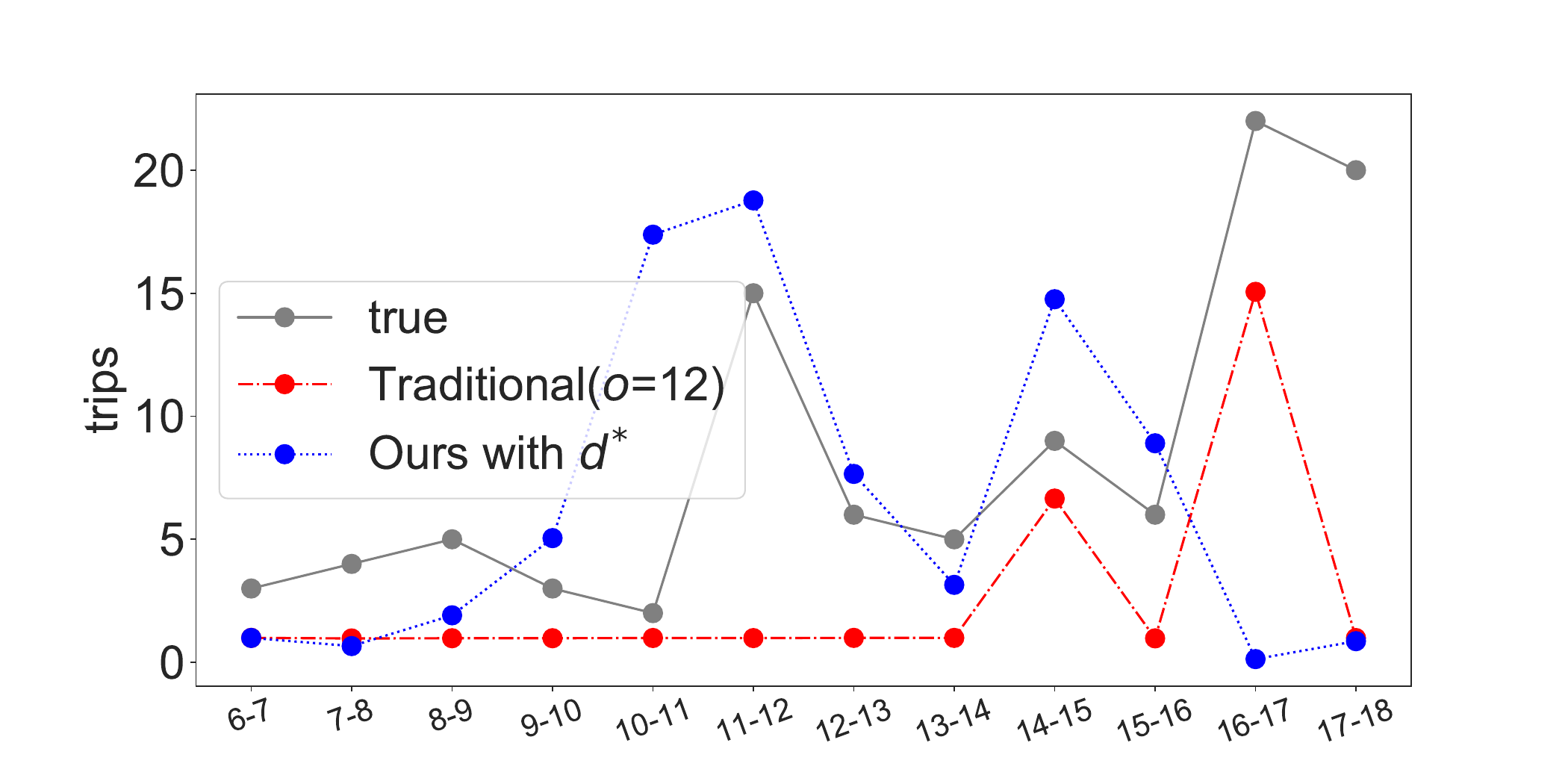}}

\caption{Time numerical evolution of different orders of magnitude (a)ODpair1-10 (b)ODpair10-8 (c)ODpair0-14 (d)ODpair12-3 } 
\label{} 
\end{figure*}

\begin{figure*}
\centering
\subfigure[OD8-9]{\includegraphics[width=4.4cm]{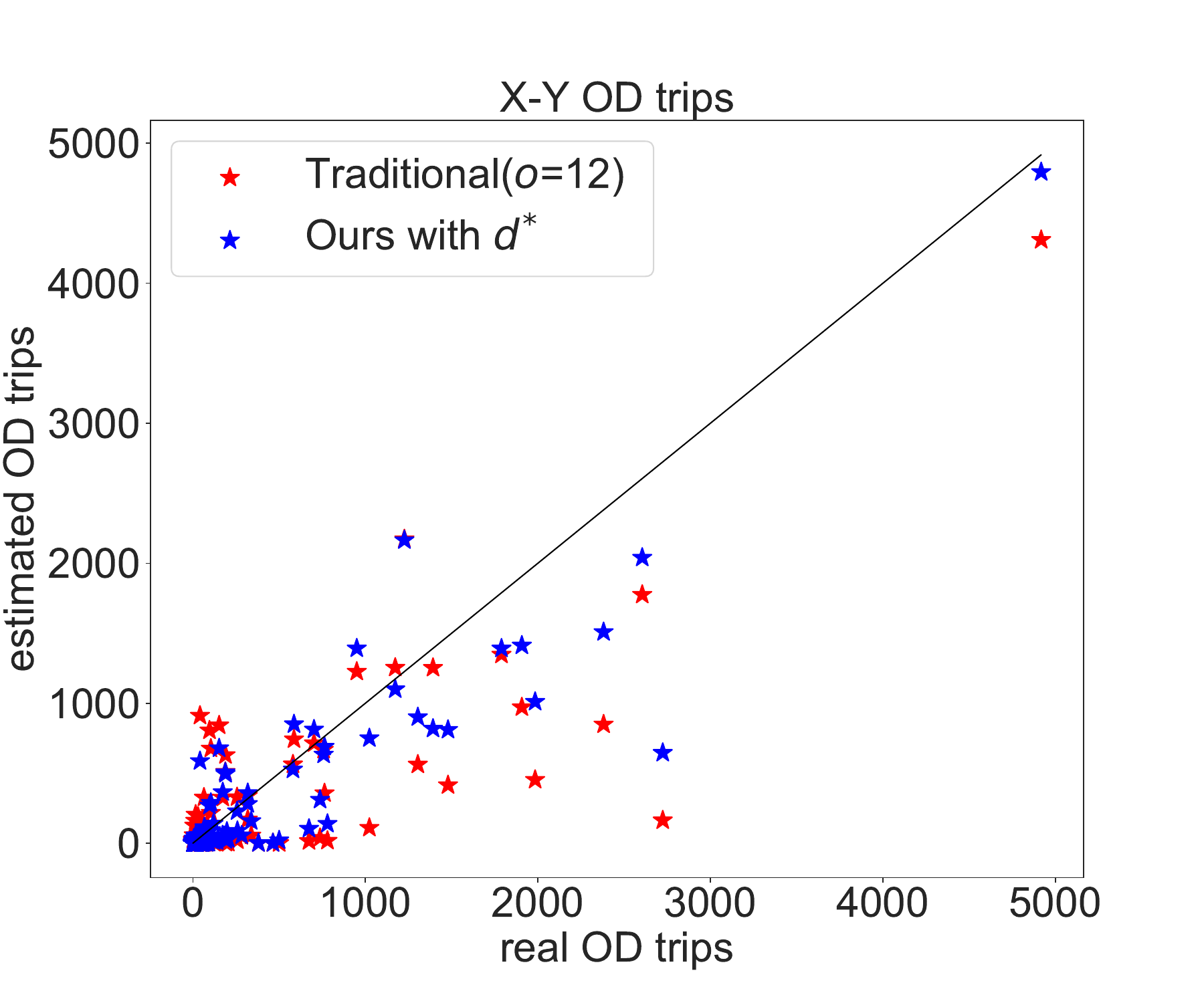}}
\subfigure[OD9-10]{\includegraphics[width=4.4cm]{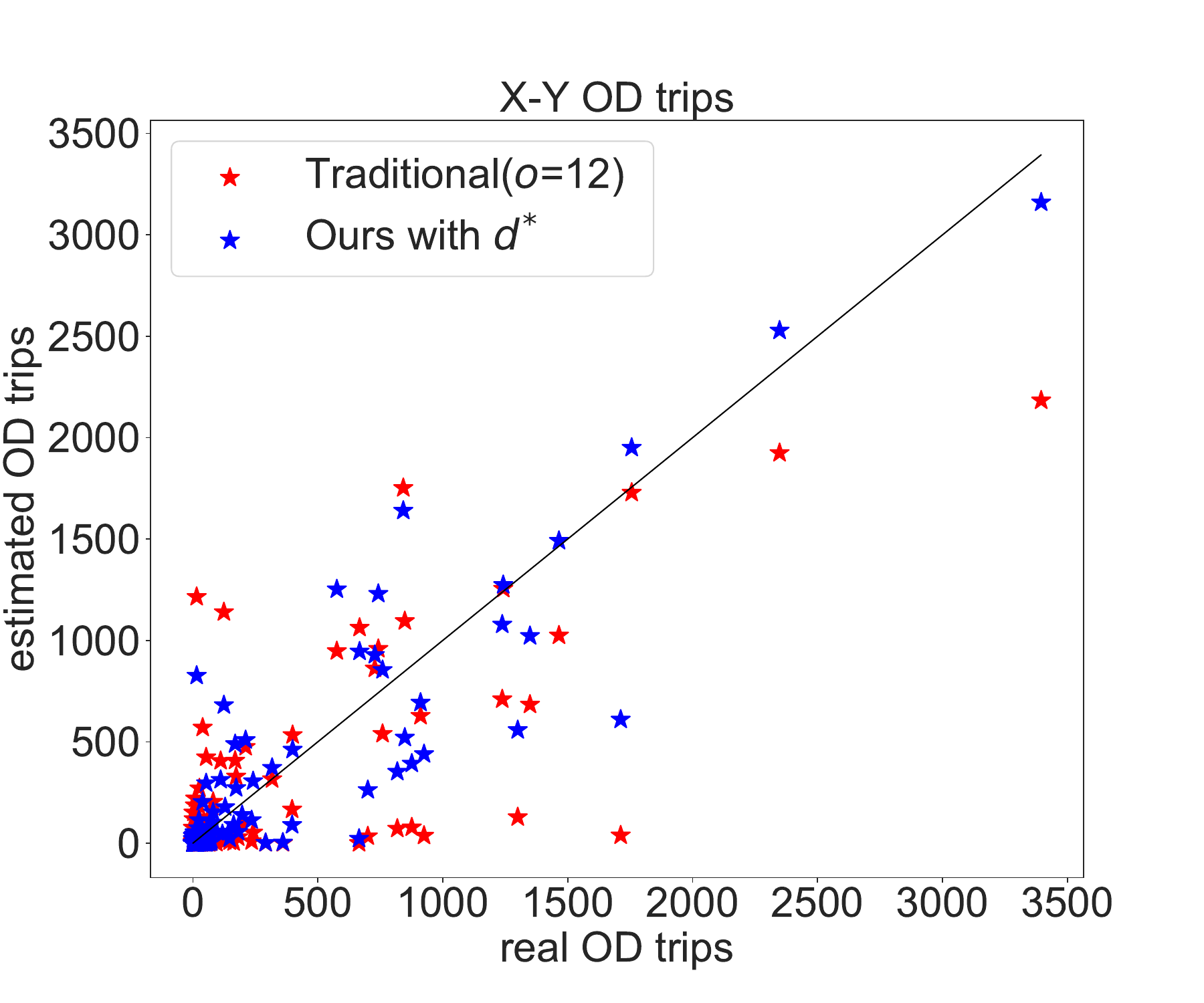}}
\subfigure[OD12-13]{\includegraphics[width=4.4cm]{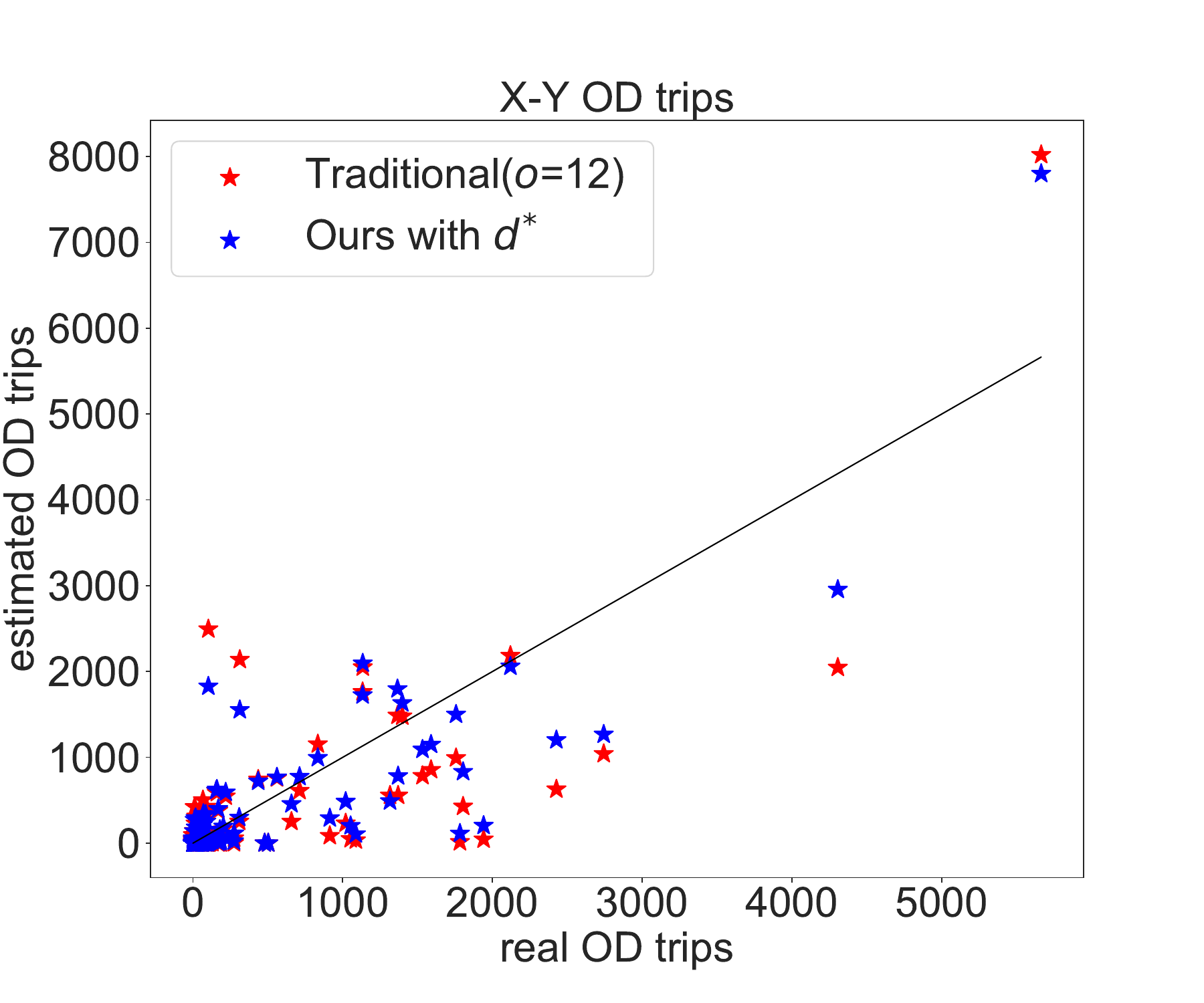}}
\subfigure[OD14-15]{\includegraphics[width=4.4cm]{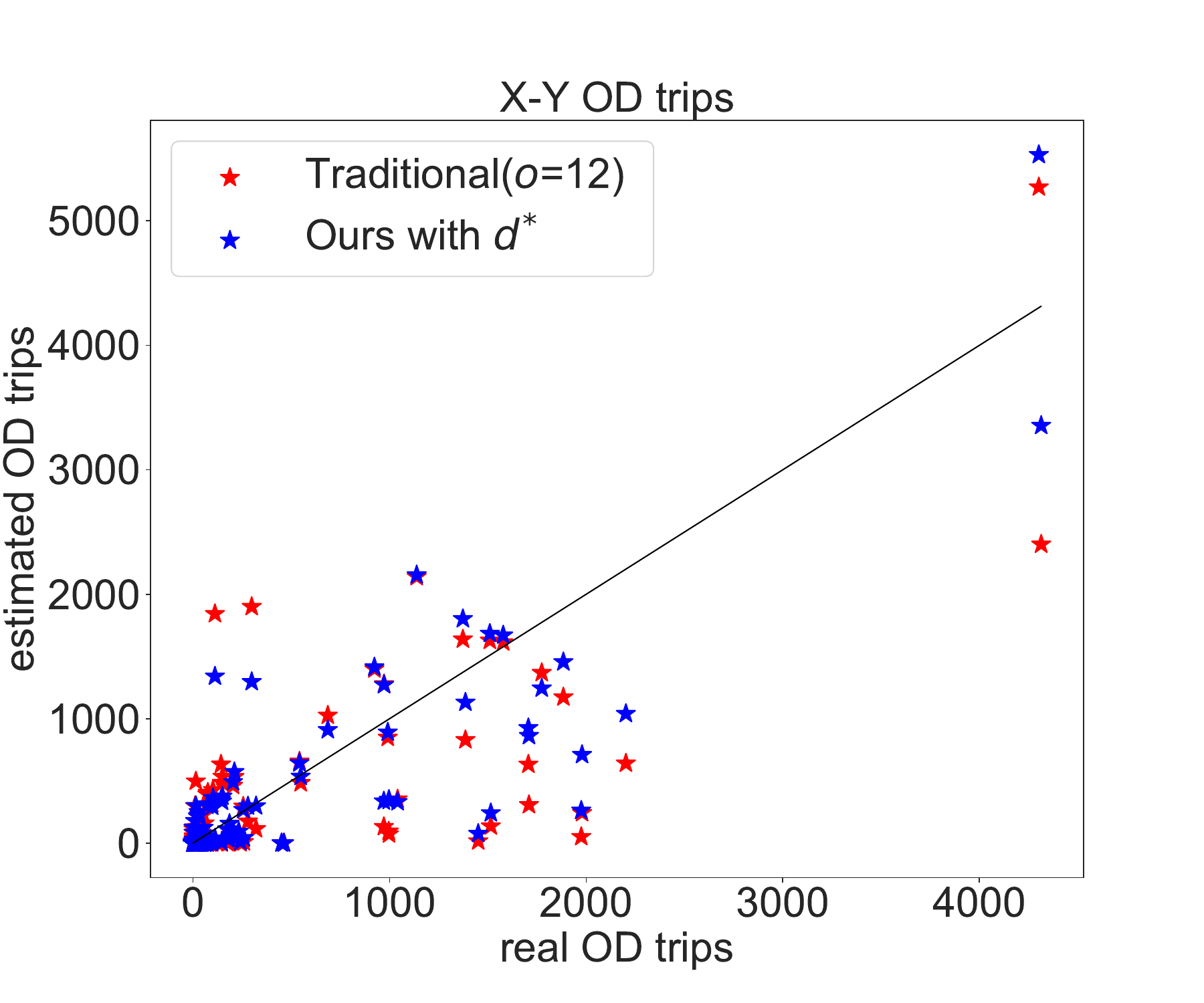}}

\caption{x-y} 
\label{} 
\end{figure*}

\section{CONCLUTION}
In this paper, we propose a deep learning method to learn the relationship between traffic counts and OD structure information through mining information from a small amount of mixed observational traffic data. Then we use this structure information to constrain the traditional least squares numerical optimization method based on the bi-level framework. We validate that our method outperforms traditional numerical-only optimization methods on 12 hours of synthetic data from a large-scale city. Moreover, we present the space for future improvement of our method by improving the sampling method and deep learning models.

\bibliography{A_DeepLearning_Framework_for_Dynamic_Estimation_of_Origin_Destination_Sequence}
\bibliographystyle{IEEEtran}

\end{document}